\documentclass[10pt,twocolumn,letterpaper]{article}

\usepackage{iccv}              

\makeatletter
\renewcommand{\paragraph}{%
  \@startsection{paragraph}{4}%
  {\z@}{0.8ex \@plus 0.2ex \@minus .2ex}{-1em}%
  {\normalfont\normalsize\bfseries}%
}
\makeatother

\definecolor{iccvblue}{rgb}{0.21,0.49,0.74}
\usepackage[pagebackref,breaklinks,colorlinks,allcolors=iccvblue]{hyperref}
\usepackage{adjustbox}
\usepackage{subcaption}
\def\paperID{2394} 
\def\confName{ICCV}
\def\confYear{2025}

\title{Better Together: Unified Motion Capture and 3D Avatar Reconstruction}

\author{Arthur Moreau$^{1}$
\quad Mohammed Brahimi$^{1,2}$ \quad Richard Shaw$^{1}$ \quad Athanasios Papaioannou$^{1}$ \\
Thomas Tanay$^{1}$ \quad Zhensong Zhang$^{1}$ \quad Eduardo Pérez-Pellitero$^{1}$
\\ \textsuperscript{1}Huawei Noah's Ark Lab \textsuperscript{2} TU Munich}

\begin{document}
\maketitle
\begin{abstract}
We present \textit{Better Together}, a method that simultaneously solves the human pose estimation problem while reconstructing a photorealistic 3D human avatar from multi-view videos. While prior art usually solves these problems separately, we argue that joint optimization of skeletal motion with a 3D renderable body model brings synergistic effects, i.e. yields more precise motion capture and improved visual quality of real-time rendering of avatars. To achieve this, we introduce a novel animatable avatar with 3D Gaussians rigged on a personalized mesh and propose to optimize the motion sequence with time-dependent MLPs that provide accurate and temporally consistent pose estimates. We first evaluate our method on highly challenging yoga poses and demonstrate state-of-the-art accuracy on multi-view human pose estimation, reducing error by 35\% on body joints and 45\% on hand joints compared to keypoint-based methods. At the same time, our method significantly boosts the visual quality of animatable avatars (+2dB PSNR on novel view synthesis) on diverse challenging subjects.

\end{abstract}    
\section{Introduction}
\label{sec:intro}

Capturing the intricate nuances and subtle details of the human body and its movements from images is a challenging task due to the complex interplay of geometry, anatomy, and dynamic motion. By decoupling these problems into distinct sub-tasks, significant progress has been made over the past decade. Given images or videos of a human subject, markerless motion capture methods~\cite{desmarais2021review, stathopoulos2024score, hmrKanazawa17}, are now capable of estimating the 3D skeleton pose of the subject through regression or optimization of keypoint alignment, enabling multiple downstream applications in virtual content creation~\cite{menolotto2020motion}. One such application is the reconstruction of controllable 3D virtual avatars of the captured subject~\cite{moreau2024human}, which have recently reached a high degree of realism thanks to novel radiance field representations~\cite{nerf,kerbl3Dgaussians}. 

\begin{figure}[ht]
\centering
\includegraphics[width=\columnwidth]{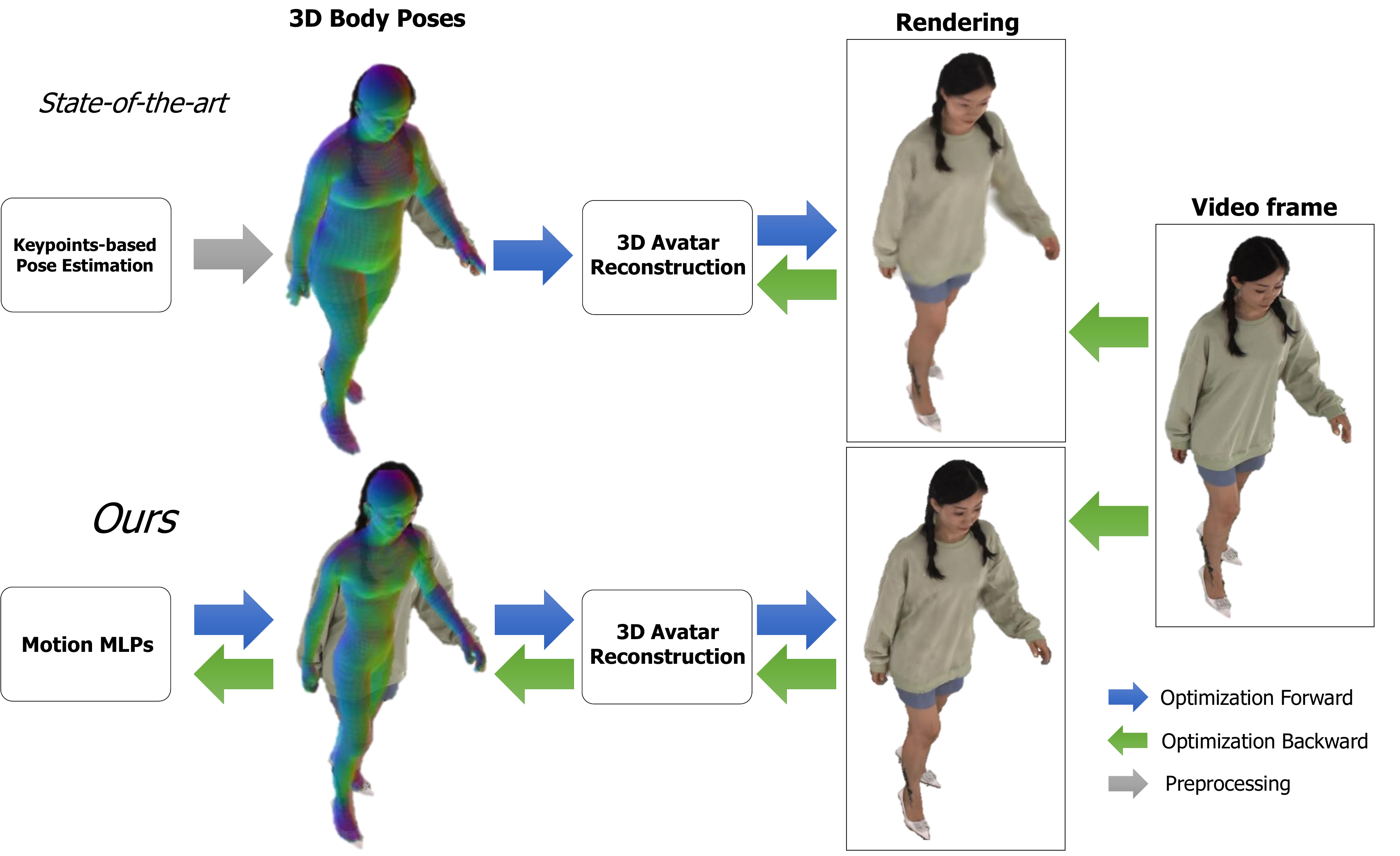}
\caption{Our method iteratively optimizes human poses with an animatable avatar to reconstruct images. We observe that photometric supervision enables to learn not only photorealistic avatars, but also highly accurate human motion.}
\label{fig:teaser_figure}
\end{figure}

However, as shown in Figure~\ref{fig:teaser_figure}, inaccuracies in the motion capture process result in reduced visual quality for the avatars because the deformed skeleton is misaligned with the image. On the other hand, existing pose estimation methods operate by optimizing keypoints position. While keypoints are a strong signal to robustly guide the skeleton toward the approximate joint location, they lack accuracy because determining the location of a body joint in an image is ambiguous since the skeleton is not directly observable. The purpose of body joints is to model the articulations of the human body, and we believe that the best way to infer them is to recover the positions that enable to explain the motion observed in a video, rather than using detection.

In this work, instead of treating human pose estimation (HPE) and avatar creation in separate consecutive blocks, we propose to solve them jointly in a pipeline that optimize the reconstruction of images, hoping for synergistic effects. Our algorithm can be seen as a markerless motion capture system that uses a more realistic body model than previous work, enabling fine-grained pose estimation at a pixel level, but also as an improved avatar reconstruction pipeline that does not require pre-computed SMPLx parameters. 

In a more formal way, we make the hypothesis that photometric alignment of a renderable body reconstruction can provide a more precise and dense supervision signal than keypoints for 3D human pose estimation. If a 3D body model is deformed, rendered and optimized to align with pixels from observed images, the only way it has to solve the problem is to estimate the correct skeleton motion, and our goal is to learn it this way. Testing this idea involves having a differentiable model capable of rendering a subject in any pose in a photorealistic way. To achieve this, we propose a novel animatable body model equipped with 3D Gaussians~\cite{kerbl3Dgaussians} for fast and high-quality rendering. Gaussian primitives are defined locally on triangles of a personalized mesh that fits the body shape of the subject. Body poses (including articulated jaw and fingers) are optimized using a time-dependent neural network that provides accurate and smooth human motion. All these modules are learned together in a differentiable pipeline supervised with images. Our contributions can be summarized as follows:
\begin{enumerate}
    \item A state-of-the-art motion capture algorithm that leverages dense photometric supervision on multi-view videos (starting from 2 cameras).
    \item A novel local parametrization of 3D Gaussians on mesh that improves the quality of animatable human models.
    \item The use of time-dependent MLPs to fit the motion sequence more precisely.
\end{enumerate}
\section{Related work}
\label{sec:related_work}

\paragraph{3D Reconstruction with Radiance Fields}

In the past few years, the field of neural rendering has progressed significantly with the development of 3D scene representations learned from 2D images, and able to perform (differentiable) novel view synthesis in a photorealistic way, that we refer to as radiance fields. It started with NeRF~\cite{nerf} and its further improvements~\cite{mueller2022instant, barron2022mipnerf360, barron2023zipnerf}, which use a neural scene representation combined with volumetric rendering. Recently, 3D Gaussian Splatting (3DGS)~\cite{kerbl3Dgaussians} addressed the same task with a different approach: the scene is represented by a collection of primitives shaped as 3D Gaussians. Rendering is performed by tile-based rasterization, which is significantly more efficient than volumetric rendering used in NeRFs. Radiance fields have been extended to model dynamic scenes~\cite{li2020neural, park2021nerfies, Wu_2024_CVPR, shawswings} and thus can be used to represent video data with humans in motion.

While radiance fields are originally designed to generate photorealistic images, the differentiability of their rendering operation can be used to optimize various parameters of the 3D scene in a render-and-compare approach. This technique provides precise results because it optimizes the photometric alignment between the rendered image and the groundtruth at a pixel level. The idea has been exploited to address many problems, such as point tracking~\cite{seidenschwarz2024dynomo}, camera pose estimation~\cite{bian2023nope, Fu_2024_CVPR}, SLAM~\cite{rosinol2023nerf, Matsuki:Murai:etal:CVPR2024}, camera relocalization~\cite{yen2020inerf, lens, moreau2023crossfire}, object pose estimation~\cite{Guo2024ECCV, cai_2024_GSPose},  and sensor calibration~\cite{herau2023moisst, herau2024soac, herau20243dgs}. Our work explores its use for 3D human pose estimation.

\paragraph{Human Body Models}

To represent the body shape of humans, an important milestone has been the creation of statistical mesh templates such as SMPL~\cite{SMPL:2015} and SMPLx~\cite{SMPL-X:2019}. They define a 3D mesh parametrized by a set of low-dimensional parameters representing variations in body shape and pose. These templates are fitted on large datasets of 3D scans representing diverse humans, such that body shape parameters can represent a large variety of body types. Textured meshes are the most common representation of 3D avatars~\cite{henderson2020leveraging, lin2021mesh, chen2024meshavatar}. While they are heavily optimized and convenient to use, their ability to render photorealistic humans is limited. Building animatable body models with radiance fields has been a very active research area. Body pose control is not easy to combine with NeRF, but has been achieved by several works~\cite{peng2021animatable, peng2021neural, li2022tava, SLRF}. In contrast, the explicit primitives proposed by 3DGS~\cite{kerbl3Dgaussians} enables easier deformation of the model and real-time rendering, leading to numerous proposals of Gaussian-based human body models~\cite{moreau2024human, wen2024gomavatar, li2024animatable, hu2024gaussianavatar, qian20243dgs, pang2024ash, PhysAvatar24, li2024gaussianbody, jiang2024hifi4g, dhamo2023headgas}.
Human Gaussian Splatting (HuGS)~\cite{moreau2024human} showed that Gaussian primitives can be directly deformed with linear blend skinning and non-rigid motion, similar to mesh vertices. Another popular approach has been to maintain an articulated mesh representation to represent geometry and define Gaussians locally on triangle meshes~\cite{wen2024gomavatar, qian2024gaussianavatars, PhysAvatar24}. Our work also builds on this direction and proposes a novel parametrization of mesh and Gaussians with subject-specific MLPs.

\paragraph{Human Pose Estimation}

The goal of HPE is to estimate the body joints position defined on a human skeleton. It can be done in 2D where the outputs are pixel coordinates and for which keypoint detection methods are well-established solutions~\cite{openpose, xu2022vitpose, khirodkar2025sapiens}. When multiple calibrated cameras are available, 3D pose estimation can be performed by triangulation of 2D skeletons~\cite{iskakov2019learnable,remelli2020lightweight,nogueira2024markerless} into 3D keypoints, or directly from image features with end-to-end approaches~\cite{dong2019fast,tu2020voxelpose,zhang2021direct,choudhury2023tempo,liao2024multiple}. 
SMPL models have been widely adopted and thus, many pose estimation methods have focused on SMPL parameter estimation from images (skeleton pose and body shape), also known as Human Mesh Recovery (HMR). Most popular HMR solutions are regressors operating on a single image~\cite{hmrKanazawa17,xiang2019monocular,goel2023humans,kocabas2024pace, dwivedi2024poco,Dwivedi_2024_CVPR}. These methods are robust enough to be used \textit{in-the-wild} and are progressing rapidly, but the resulting human motion is often imprecise and temporally inconsistent. One solution given by ProHMR~\cite{kolotouros2019spin} and ScoreHMR~\cite{stathopoulos2024score} is to perform test-time optimization with temporal or multi-view constraints to improve the quality of regressors.  When higher accuracy is needed, one can use offline optimization methods based on keypoints. In this scenario, SMPLify-X~\cite{SMPL-X:2019} and related methods~\cite{MuVS:3DV:2017,easymocap} optimize body pose and shape to align the template with keypoints detected in the input images. This is compatible with monocular (2D keypoints) or multi-view (3D keypoints) data captures.
Our work aims to improve the accuracy of these methods by rendering a Gaussian-based body model. Perhaps surprisingly, this idea has been explored back in 2011~\cite{stoll2011fast}, but the body model was built with a small number of primitives that do not provide photorealistic rendering. More recently, several Gaussian-based avatar methods include in their pipeline a pose refinement module, suggesting the capability of these models to be used for motion capture. AvatarPose~\cite{lu2024avatarpose} combines pose estimation with NeRF but their method focuses on contact refinement during multi-person interactions. All these methods refine pre-computed pose parameters whereas our method estimates them from scratch.
\section{Method}
\label{sec:method}

\begin{figure*}[h]
\centering
\includegraphics[width=0.99\textwidth]{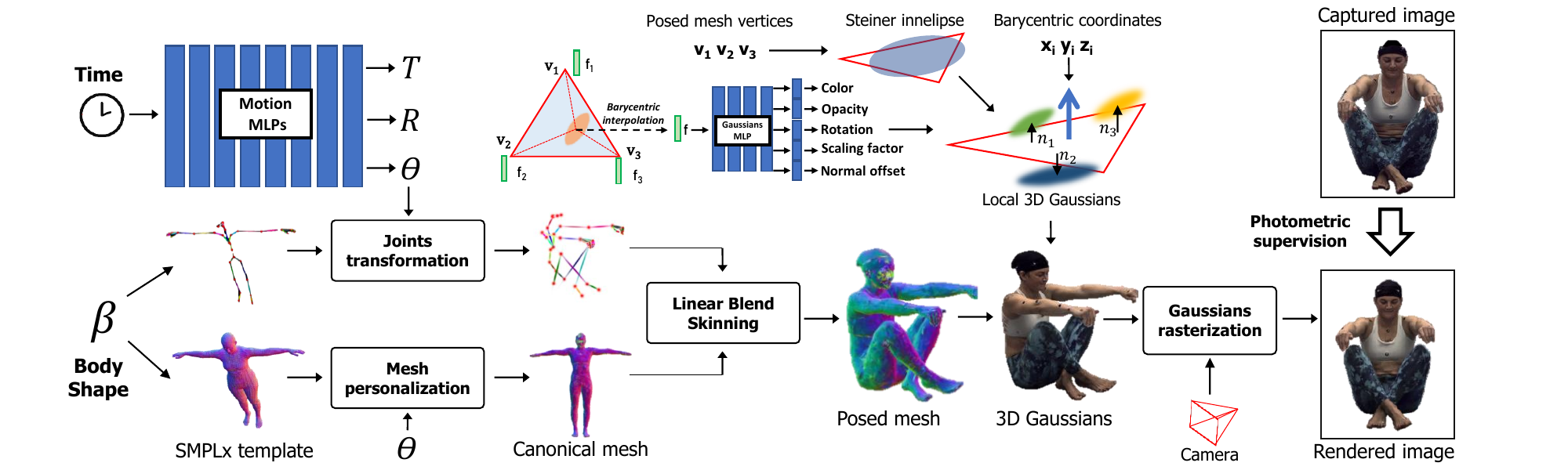}
\caption{\textbf{Overview of the forward deformation process of our method.} Given a timestep, Motion MLPs process pose parameters, which are used to compute skeleton joints transformations. Then, we compute pose-dependent personalization of the template mesh, that we deform with LBS. 3D Gaussians are then attached to triangles and rasterized into an image, which is compared to the captured frame.}
\label{fig:overview}
\end{figure*}

We propose a markerless motion capture (3D Human Pose Estimation) algorithm using multi-view RGB videos that also outputs an animatable avatar of the subject. The core principle is to optimize the alignment between the captured images and a rendering of a human body model through 3D Gaussian Splatting. We present our body model in Sec~\ref{sec:method_avatar}, our implicit motion parametrization in Sec~\ref{sec:method_motion_param}, and the procedure to optimize them together in Sec~\ref{sec:method_optimization}.

\subsection{Human Body Model}
\label{sec:method_avatar}

This section describes our body model, that produces 3D Gaussians given a target body pose $\theta$. It consists in 2 modules, one personalized mesh that carries the animation and defines the human geometry and a 3D Gaussians layer that defines renderables primitives on the mesh surface.

\subsubsection{Personalized Template Mesh}
\label{sec:method_mesh}

\paragraph{SMPLx Template Model}
Our body model is driven by the skeleton of SMPLx~\cite{SMPL-X:2019}. It consists of 55 body joints, including articulated hands and face. Body joints are linked together by bones defined in a kinematic tree.  Animation is achieved by assigning a 3D rotation to each joint, known as the body pose, and then applying these transformations sequentially along the kinematic tree. It also include a mesh template, defined by V=10475 vertices. Vertices positions $v_{temp} \in \mathbb{R}^{V \times 3}$ are controlled by body shape parameters $\beta$ that can be adjusted to fit the subject body. Because body joints are regressed from $v_{temp}$, $\beta$ also controls the shape of the skeleton (bones length). 

\paragraph{Per-vertex Offsets} SMPLx does not model clothing and its surface is never exactly aligned with the actual body surface. Our method personalizes the shape of the canonical template mesh by learning subject-specific per-vertex additive offsets $\Delta v \in \mathbb{R}^{V \times 3}$. We aim for this personalized mesh to lie as close as possible to the observed surface because the 3D Gaussians used for rendering are defined on the mesh triangles.

\paragraph{Per-vertex Latent Features} We use latent feature vectors $f \in \mathbb{R}^{V \times D}$ defined on each mesh vertex. These features, initialized randomly and then optimized, are used to store latent local information decoded by MLPs that process mesh and Gaussians parameters. Latent features are not only defined at the vertex positions, but can be queried anywhere on the mesh surface by barycentric interpolation. This design is related to explicit features grids that enhance/accelerate neural scene representations~\cite{mueller2022instant} : instead of describing the whole 3D space using hash grids or triplanes, our feature grid is defined continuously along the 2.5D human surface.

\paragraph{Mesh MLP: Pose-dependent Deformation and Shading} 
The rigid deformation provided by skeleton animation is not sufficient to explain the detailed motion observed on images, such as clothing wrinkles and muscles contraction. We use a MLP to predict the per-vertex non-rigid motion $m \in \mathbb{R}^{V \times 3}$, as well as a shading component $s \in [0,2]^V$. The network processes each vertex $i$ independently through the batch dimension and takes as input the latent feature $f_{i}$ as well as the body pose $\theta$, encoded to avoid spurious correlation following SCANimate~\cite{Saito:CVPR:2021}. We use an encoder of 4 layers with ReLU activations and both outputs are decoded by dedicated heads containing 1 hidden layer. The pose-dependent shading $s$ captures local lighting variations observed in the data and serves as a multiplicative factor on the RGB color of each Gaussian.

The final positions of canonical vertices are given by:
\begin{equation}
    v = v_{temp}(\beta) + \Delta v + m(\theta)
\end{equation}

We manually set $\Delta v = 0$ and $m(\theta) = 0$ for hands and face vertices, where we observe artifacts if personalized.

\paragraph{Animation} The canonical mesh is deformed to the target body pose via forward Linear Blend Skinning, using skinning weights from the SMPLx template.

\subsubsection{3D Gaussians Layer}
\label{sec:gaussians_layer}

Each Gaussian primitive of our model is attached rigidly to a parent triangle from the personalized mesh. This idea has been recently explored by several works~\cite{guedon2024frosting,wen2024gomavatar, qian2024gaussianavatars, PhysAvatar24}. It enables to leverage the animation capabilities of the mesh and thus to build controllable photorealistic models.

\paragraph{Gaussians Shaped as Steiner Ellipses} The Steiner ellipse of a triangle is the only ellipse that intersects the 3 vertices and whose center is the triangle centroid. The Steiner inellipse is similar but downscaled by a factor 2 and intersects triangle mid-points. In other words, Steiner ellipses are optimal to approximate the triangle shape with an ellipse. Their parameters can be computed differentiably from vertex positions. We see this as an opportunity to render meshes with Gaussian rasterization. We define a \textit{Steiner Gaussian} primitive which is the 3D extension of 2D Steiner ellipses. The scaling parameter along the additional dimension, defined along the triangle's normal direction is set to a constant $\epsilon = 1e-5$ to define a flat primitive lying on the triangle surface. Because the primitive parameters depend on vertex positions, when a triangle is deformed during animation, the shape of our primitive is deformed accordingly to approximate the triangle. By defining one Steiner Gaussian per triangle, we obtain a real-time and differentiable (approximate) mesh renderer. Mesh visualizations shown in Figure~\ref{fig:quali_moyo} are rendered with Steiner Gaussians color-coded with their normal direction. However, we observe that using more than one Gaussian per triangle enables better fitting of details on the subject appearance. 

\paragraph{Barycentric Coordinates} To define multiple Gaussian primitives per triangle, we equip them with barycentric coordinates that express the position of any point located on a triangle. We learn 3 floats per primitive (initialized with ones), which are processed with a softplus activation and divided by their sum to obtain barycentric coordinates $x_{i},y_{i},z_{i}$ that sum to 1.

\paragraph{Implicit Parametrization of Gaussians} To obtain Gaussian parameters, we first compute the Steiner Gaussian of each triangle. Each primitive is defined as a deformed version of the Steiner Gaussian by using 3 scaling multiplicative factors between 0 and 1 (1 being the scale of the Steiner Gaussian) and a 3D orientation applied to rotate the Steiner Gaussian. Opacity, used for alpha-blending of primitives, is defined between 0 and 1. We do not use a view-dependent color formulation, but shading values obtained from the mesh provide pose-dependent color variations. We also introduce an additional scalar parameter, the normal offset $o_{i}$, that allows each Gaussian to slightly deviate from the mesh surface to fit the 3D geometry with more precision.    
To obtain parameters, we perform barycentric interpolation of  $f_{i}$ to obtain one latent feature for each Gaussian position on the surface. This vector is the input to a 4-layer MLP with ReLU activations that outputs the parameters through dedicated heads with one hidden layer. The center position of a Gaussian primitive  $i$ defined in the triangle of vertices $v_1, v_2, v_3$ with normal direction $\vec{n}$  is given by:
\begin{equation}
    c_{i} = x_{i} v_1 + y_i v_2 + z_i v_3 + o_i \vec{n} 
\end{equation}

\subsection{Human Motion Parametrization}
\label{sec:method_motion_param}

Our goal is to learn accurately the 3D skeletal motion of the subject, i.e. joints orientation $\theta_{t}$ and global translation $T_{t}$ and orientation $R_{t}$ for each timestep $t$ of the video.

\paragraph{Motion MLPs} Similar to~\cite{he2022nemf} we use neural networks to fit human motion over time. Such a design enables to leverage the smoothness bias of neural networks to obtain consistent motion trajectories over time. We optimize 3 separate MLPs that output respectively body pose parameters, hand pose parameters, and global transformation, that we observe to perform better than a single shared network. Body and hands MLPs have 8 fully-connected layers with ReLU activations, while the network responsible for global transformation has 4 layers. Body pose parameters and global orientation are predicted directly in the axis-angle representation. In Sec~\ref{sec:ablations}, we compare our solution with a discrete representation and with the use of a human body prior.

\paragraph{Anatomy consistent hand pose parameters} SMPLx uses MANO~\cite{MANO:SIGGRAPHASIA:2017} for the hand skeleton. It consists of 3 body joints per finger, resulting in 45 parameters per hand in the axis-angle representation. However, hand articulation presents effectively much less DoF such that predicting these parameters directly leads to unrealistic hand poses. MANO provides a PCA encoded pose representation with 6DoF that we observe to not be expressive enough. Instead, we use a customized parameters space with 22DoF per hand that constraints finger poses to anatomically plausible movements, related to CPF~\cite{yang2021cpf}. Implementation details are given in supplementary materials.

\subsection{Optimization Process}
\label{sec:method_optimization}

The explicit goal of our method is to reconstruct images $I_{t,c}$ captured at timestep $t$ from camera $c$. At each iteration, we use one batch of images captured at different timesteps from the same camera, deform the body model, render images as described in Figure~\ref{fig:overview}, and backpropagate the gradients back to the learnable parameters. These parameters are: the body shape $\beta$, per-vertex offsets $\Delta v$ and latent features $f$, per-Gaussian barycentric coordinates $x,y,z$ and the weights of the networks that learn respectively pose, mesh and Gaussian parameters.

\textbf{Incremental Training.} Photometric losses have a small convergence basin and thus cannot guide the convergence of the pose network when the body is far away from the correct solution. We address this problem by first training the pose network to align the SMPLx template model with 3D keypoints without rendering images. This pretraining step is efficient because discarding images enables to train with full batch (the body model is deformed for each timestep of the video at each iteration). It can be trained in 5 minutes and provides reasonable body poses. During this step only, we add a regularization term on the norm of pose parameters, to avoid unrealistic twists of joints. Then, we continue to optimize the motion MLPs jointly with the Gaussian body model to reconstruct images.

\textbf{Image Reconstruction Losses.} Our optimization objective combines the three most popular photometric losses, namely L1, SSIM~\cite{wang2004image} and LPIPS~\cite{zhang2018unreasonable}. To preserve efficiency, LPIPS is not computed on full images, but on 2 randomly sampled patches of $64 \times 64$ pixels. These patches are located around projected body joints, such that relevant parts of the image are used. 

\textbf{Mesh normals regularization:} To ensure a regular geometry of the learned mesh, we encourage the normal directions of pairs of neighouring faces $\mathbf{n}_i$ and $\mathbf{n}_j$ to be similar. \[
\mathcal{L}_{\text{normals}} = \sum_{(i,j) \in \mathcal{E}} \left( 1 - \mathbf{n}_i \cdot \mathbf{n}_j \right)
\]

\textbf{Mesh MLP regularization:} We regularize the mesh MLP by minimizing the impact of his outputs (\ie small non-rigid motion and color variation).
\[\mathcal{L}_{\text{meshMLP}} = \frac{1}{V} \sum_{i=1}^V \lambda_{m} |m_i| + \lambda_{s}|s_i - 1|
\]

\textbf{Gaussian regularization
:} We minimize the value of Gaussian normals offset $o_{i}$, first to prevent Gaussians from operating far from the mesh surface, but also to push the mesh faces close to the actual surface fitted by Gaussians. We also penalize the rotational offset of Gaussians for angles that deviate from the normal direction, such that we preserve a smooth envelope of Gaussians. The rotation residual provided by the Gaussians MLP is parametrized with Euler angles $(\alpha,\beta,\gamma)$. $\alpha$ rotates along the normal direction and is free, but we penalize $\beta,\gamma$.
\[
\mathcal{L}_{\text{Gaussians}} = \frac{1}{G} \sum_{i=1}^G \lambda_o o_i^2 + \lambda_{rot}(\beta_i^2 + \gamma_i^2)
\]

\paragraph{Implementation details} \textit{Better Together} is implemented in the Pytorch~\cite{paszke2019pytorch} framework. We use the Gaussian rasterizer released by the original 3DGS implementation~\cite{kerbl3Dgaussians}. We optimize for 5k iterations in the pretraining step and 40k with photometric supervision with a batch size of 32. The training takes approximately 4 hours on a NVIDIA V100 GPU. More details are given in supplementary materials.
\begin{table*}[h]

\caption{Body and Hands Pose Estimation evaluation results on MOYO (millimeters, lower is better).}
\label{tab:pose}
\begin{adjustbox}{width=\textwidth}
\begin{tabular}{lllllllllllllllll}
\toprule
\multicolumn{2}{c}{\textbf{Body Pose Estimation}} & \multicolumn{3}{c}{\textbf{Boat Pose}} & \multicolumn{3}{c}{\textbf{Camel Pose}} & \multicolumn{3}{c}{\textbf{Cobra Pose}} & \multicolumn{3}{c}{\textbf{Firefly Pose}} & \multicolumn{3}{c}{\textbf{Average}} \\
  \cmidrule(r){3-5} \cmidrule(r){6-8} \cmidrule(r){9-11} \cmidrule(r){12-14} \cmidrule(r){15-17} 
\textbf{Method}& \textbf{Keypoints} & \footnotesize{\textit{W-MPJPE}} & \footnotesize{\textit{MPJPE}} & \footnotesize{\textit{PA-MPJPE}} & \footnotesize{\textit{W-MPJPE}} & \footnotesize{\textit{MPJPE}} & \footnotesize{\textit{PA-MPJPE}} & \footnotesize{\textit{W-MPJPE}} & \footnotesize{\textit{MPJPE}} & \footnotesize{\textit{PA-MPJPE}} & \footnotesize{\textit{W-MPJPE}} & \footnotesize{\textit{MPJPE}} & \footnotesize{\textit{PA-MPJPE}} & \footnotesize{\textit{W-MPJPE}} & \footnotesize{\textit{MPJPE}} & \footnotesize{\textit{PA-MPJPE}} \\
\cmidrule(r){1-2} \cmidrule(r){3-5} \cmidrule(r){6-8} \cmidrule(r){9-11} \cmidrule(r){12-14} \cmidrule(r){15-17} 
Ours & Sapiens & \textbf{34.30} & 45.16 & \textbf{23.77} & \textbf{34.05} & \textbf{53.71} & \textbf{28.31} & \textbf{31.03} & \textbf{35.74} & \textbf{23.13} & \textbf{31.05} & \textbf{34.31}& \textbf{23.34} & \textbf{32.61} & \textbf{42.23} & \textbf{24.64} \\
Ours (kp only) & Sapiens & 50.07 & 66.74 & 33.21 & 61.82& 89.13 & 52.09 & 38.65& 39.30 & 27.15 & 56.33 & 60.05& 38.22 & 51.72 & 63.81 & 37.67 \\
EasyMocap & Sapiens& 44.48 & \textbf{41.9} & 26.9 & 63.28 & 71.18 & 55.89 & 49.43 & 57.01 & 41.82 & 58.29 & 74.05 & 51.83 & 53.87 & 61.04 & 44.11 \\
MV-SMPLifyX & Sapiens& 59.2 & 61.13 & 37.23 & 90.09 & 120.99 & 79.85 & 63.3 & 92.48 & 50.42 & 66.42 & 93.7 & 48.37 & 69.75 & 92.08 & 53.97 \\
ScoreHMR& Sapiens& \multicolumn{1}{c}{-} & 88.52 & 44.35 & \multicolumn{1}{c}{-} & 142.93 & 67.26 & \multicolumn{1}{c}{-} & 108.38 & 50.59 & \multicolumn{1}{c}{-} & 86.31 & 62.02 & \multicolumn{1}{c}{-} & 106.54 & 56.06 \\
ScoreHMR & ViTPose& \multicolumn{1}{c}{-} & 80.26 & 42.03 & \multicolumn{1}{c}{-} & 139.96 & 58.35 & \multicolumn{1}{c}{-} & 108.51 & 50.80 & \multicolumn{1}{c}{-} & 102.51 & 74.23 & \multicolumn{1}{c}{-} & 107.81 & 56.35           \\
EasyMocap & OpenPose & 57.08 & 59.03 & 47.10 & 84.32 & 97.78 & 67.48 & 45.58 & 52.15 & 38.29 & 139.62 & 151.77 & 134.98 & 81.65 & 90.18 & 71.96 \\
\bottomrule
\end{tabular}
\end{adjustbox}

\begin{adjustbox}{width=\textwidth}
\begin{tabular}{lllllllllllllllll}
\multicolumn{2}{c}{\textbf{Hands Pose Estimation}} & \multicolumn{3}{c}{\textbf{Boat Pose}} & \multicolumn{3}{c}{\textbf{Camel Pose}} & \multicolumn{3}{c}{\textbf{Cobra Pose}} & \multicolumn{3}{c}{\textbf{Firefly Pose}} & \multicolumn{3}{c}{\textbf{Average}} \\
 \cmidrule(r){3-5} \cmidrule(r){6-8} \cmidrule(r){9-11} \cmidrule(r){12-14} \cmidrule(r){15-17} 
\textbf{Method}& \textbf{Keypoints} & \footnotesize{\textit{W-MPJPE}} & \footnotesize{\textit{MPJPE}} & \footnotesize{\textit{PA-MPJPE}} & \footnotesize{\textit{W-MPJPE}} & \footnotesize{\textit{MPJPE}} & \footnotesize{\textit{PA-MPJPE}} & \footnotesize{\textit{W-MPJPE}} & \footnotesize{\textit{MPJPE}} & \footnotesize{\textit{PA-MPJPE}} & \footnotesize{\textit{W-MPJPE}} & \footnotesize{\textit{MPJPE}} & \footnotesize{\textit{PA-MPJPE}} & \footnotesize{\textit{W-MPJPE}} & \footnotesize{\textit{MPJPE}} & \footnotesize{\textit{PA-MPJPE}} \\
\cmidrule(r){1-2} \cmidrule(r){3-5} \cmidrule(r){6-8} \cmidrule(r){9-11} \cmidrule(r){12-14} \cmidrule(r){15-17} 
Ours & Sapiens & \textbf{15.47} & \textbf{10.09} & \textbf{6.68} & \textbf{20.15} & \textbf{18.39} & 11.14 & \textbf{17.43} & \textbf{9.90} & 5.85 & \textbf{18.54} & \textbf{18.23} & 8.95 & \textbf{17.90} & \textbf{14.15} & 8.15 \\
Ours (kp only) & Sapiens & 17.39 & 13.28 & 7.20 & 56.95 & 23.94 & \textbf{8.83} & 18.30 & 13.54 & 5.49 & 37.93 & 23.88 & 8.54 & 32.64 & 18.66 & 7.52 \\
EasyMocap & Sapiens & 21.70 & 17.84 & 7.63 & 56.33 & 21.45 & 9.27 & 23.61 & 17.83 & 5.53 & 26.75 & 22.21 & \textbf{5.84} & 32.10 & 19.87 & \textbf{7.07} \\ 
EasyMocap & OpenPose & 24.34 & 17.84 & 7.67 & 75.70 & 29.08 & 9.37 & 25.85 & 20.46 & \textbf{5.30} & 281.25 & 53.98 & 11.15 & 101.79 & 30.34 & 8.37 \\
\bottomrule
\end{tabular}
\end{adjustbox}
\end{table*}

\section{Experiments}
\label{sec:experiments}

We present experiments on two datasets, one tailored for precise motion capture evaluation, the other for neural rendering of humans. We invite readers to watch the supplementary video for a better understanding of our results.

\subsection{Evaluation on MOYO Dataset}
\label{sec:experiments_mocap}

We first benchmark the human pose estimation accuracy of our method on the MOYO dataset~\cite{tripathi2023ipman}. It consists of yoga demonstrations captured by 8 synchronized RGB cameras. The subject is also equipped with detectable optical markers tracked by Vicon optical cameras, providing highly accurate ground truth SMPLx parameters. We select 4 sequences depicting challenging yoga poses and compare our method against state-of-the-art methods for multi-view HPE.

\begin{figure*}[h]
   \centering
   \includegraphics[width=\linewidth]{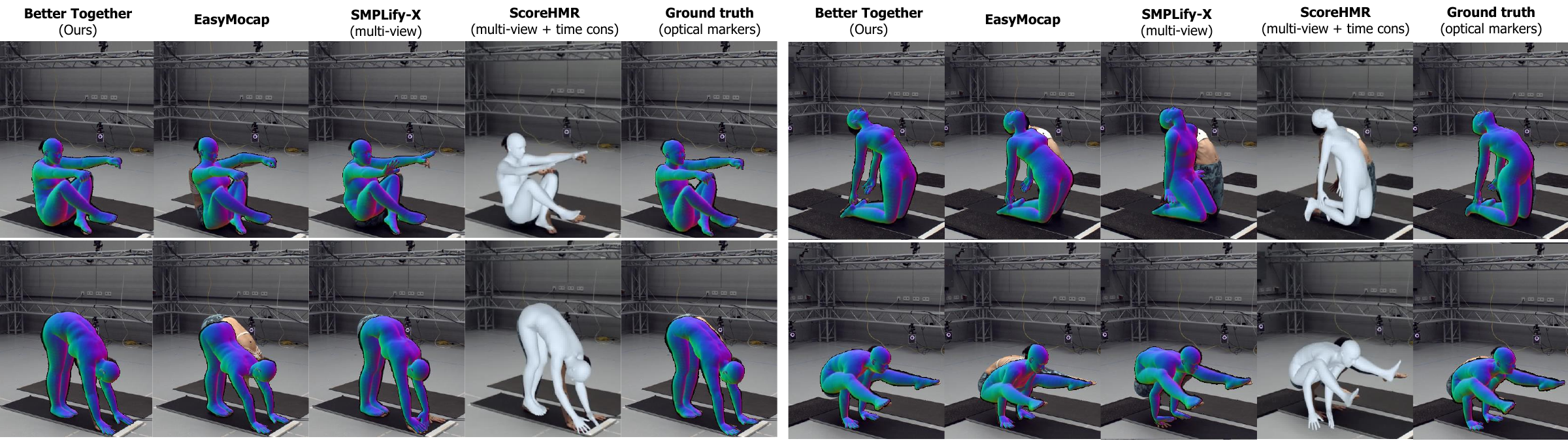}
   \caption{\textbf{Qualitative pose estimation results on MOYO.} Comparison of mesh reprojections into images. All methods use Sapiens~\cite{khirodkar2025sapiens} keypoints. Meshes are rendered with Steiner Gaussians (Sec~\ref{sec:gaussians_layer}), except ScoreHMR. Video comparison is provided in supplementary.}
   \label{fig:quali_moyo}
\end{figure*}

\paragraph{Baselines}
\begin{itemize}
    \item EasyMocap~\cite{easymocap} is the most commonly used repository for multi-view motion capture based 3D keypoints alignment. After optimization of SMPLx parameters, temporal filtering is applied to improve temporal consistency. 
    \item SMPLify-X is the optimization method proposed by SMPLx~\cite{SMPL-X:2019}, designed to align 2D keypoints with a single image. We use the multi-view implementation proposed by~\cite{zheng2020pamir}. It iteratively performs 2D keypoint alignment in each view and processes the sequence frame by frame.
    \item ScoreHMR~\cite{stathopoulos2024score} is a state-of-the-art test-time optimization method that refines predictions of a pose regressor, HMR2~\cite{goel2023humans}. It operates on single images but can incorporate multi-view and temporal constraints. We first apply multi-view consistency followed by temporal consistency, that we observe to work better than the opposite.
    \item Ours (keypoints only) is our pretraining method described in section~\ref{sec:method_optimization}. It uses our time MLP to learn body poses but only to deform the SMPLx mesh to align with 3D keypoints, similar to existing optimization approaches.   
\end{itemize}
To ensure fair comparison, we compute each baseline with both the keypoint detector recommended in their repository (OpenPose~\cite{openpose} or ViTPose~\cite{xu2022vitpose}) and the keypoints used by our method (Sapiens~\cite{khirodkar2025sapiens}).

\paragraph{Metrics} The accuracy of 3D human poses is evaluated by comparing 3D joints positions against ground truth. We use 3 variants of Mean Per Joint Projection Error (MPJPE): W-MPJPE computes the error in the world coordinates system without any post-processed alignment. We also follow common practice and report MPJPE which aligns the pelvis position and PA-MPJPE which aligns both skeletons with the Procrustes method. It should be noted that MPJPE and PA-MPJPE discard the influence of global transformation. Because some baselines do not model articulated fingers, we separate the evaluation in body pose estimation (SMPL body joints) and hands pose estimation.

\paragraph{Results}

\begin{figure*}[h]
   \centering
   \includegraphics[width=\linewidth]{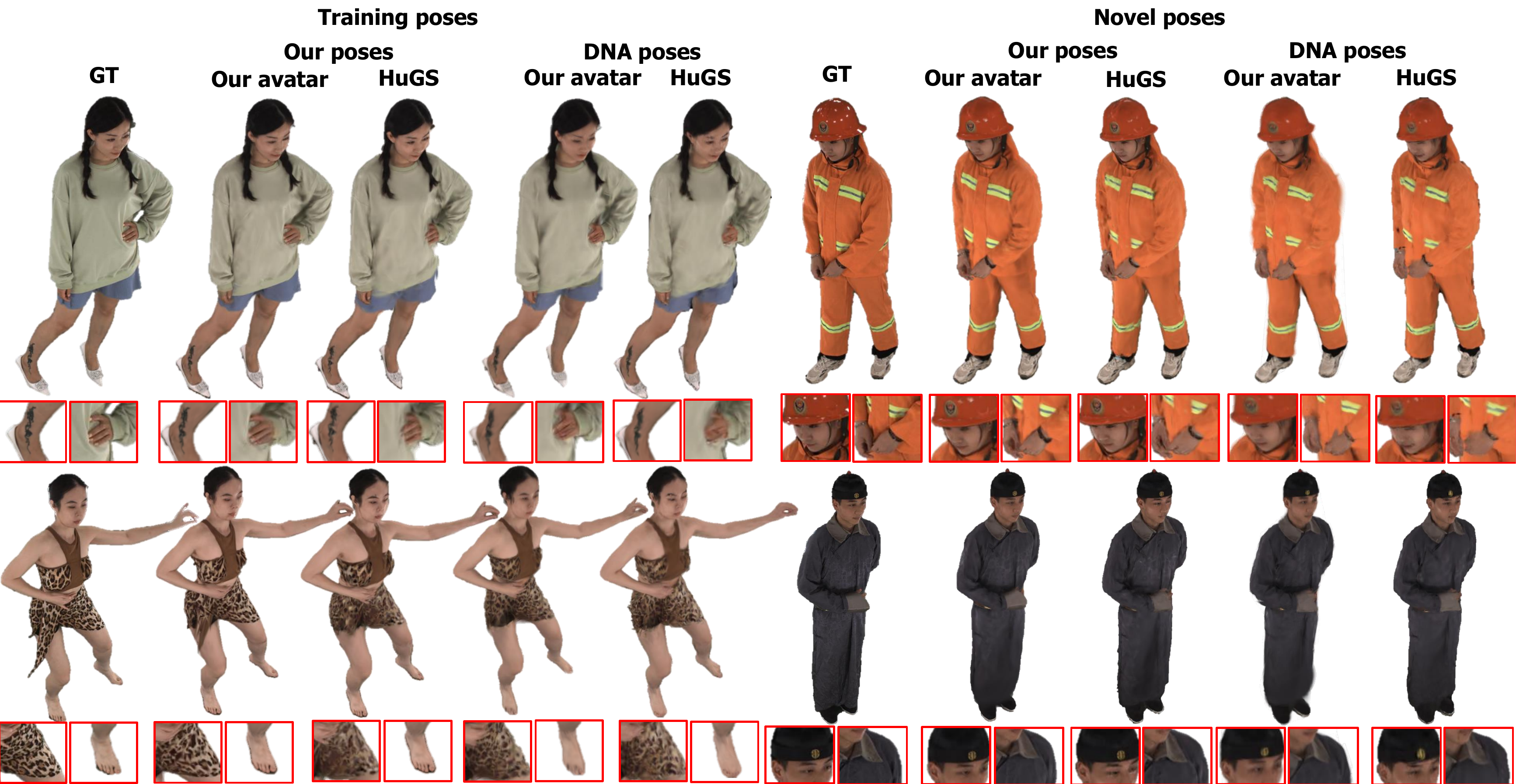}
   \caption{\textbf{Novel view synthesis on DNA Rendering.} We compare rendering quality for avatars trained with poses obtained from our method and with ground truth poses from DNA. Our parameters produce sharper rendering with details closer to GT.}
   \label{fig:quali_DNA_2}
\end{figure*}

We present pose estimation scores in Table~\ref{tab:pose}. We show a qualitative comparison of body poses in Figure~\ref{fig:quali_moyo}, and refer to the supplementary video for a complete visualization. Our method significantly outperforms baselines for both body and hands pose estimation. On the body joints, the error is reduced by around 35\% against the second best method for all metrics. For hands, W-MPJPE is improved by 45\%, meaning that our method provides way more accurate hands location. On PA-MPJPE, our method is only 1 millimeter less accurate than EasyMocap. The second best performing methods are EasyMoCap and our baseline, optimized solely on keypoints, that both present comparable results. Beyond the numbers, we observe that Better Together provides an excellent alignment when re-projected in the images thanks to the pixel-level photometric supervision. In contrast, keypoint-based methods present inaccuracies on the back area, where the information given by keypoints is sparse.   Because the regression-based method ScoreHMR does not use camera poses as input, we can not evaluate its results with W-MPJPE. This method presents good image reprojection but the poses are not 3D consistent, resulting in poor quantitative metrics. Overall, \textit{Better Together} significantly improves the state-of-the-art on this dataset.

\subsection{Evaluation on DNA-Rendering Dataset}
\label{sec:experiments_avatar}

We use DNA-Rendering dataset~\cite{cheng2023dna} to evaluate the impact of our method on avatar rendering quality. It contains multi-view captures of diverse subjects, clothing and motion. 

\begin{table}[hb]
    \centering
    \caption{Novel View Synthesis evaluation on DNA-Rendering.}
    \label{tab:DNA}
    \begin{adjustbox}{width=\columnwidth}
    \begin{tabular}{llcccccr}
    \toprule
         &  & \multicolumn{3}{c}{\textit{Training Poses}} & \multicolumn{3}{c}{\textit{Novel Poses}} \\
        \cmidrule(r){3-5}  \cmidrule(lr){6-8}
        Pose & Avatar & \textbf{PSNR} & \textbf{SSIM} & \textbf{LPIPS} & \textbf{PSNR} & \textbf{SSIM} & \textbf{LPIPS}  \\ 
        \cmidrule(r){1-5}  \cmidrule(lr){6-8}
        \multicolumn{2}{c}{\textit{Better Together}} & 29.31 & 0.9660 & 0.0654 &  \multicolumn{3}{c}{-}  \\
        \midrule
        Ours & Ours & 29.20 & 0.9654 & 0.0655 & 27.84 & 0.9587 & 0.0698  \\
        Ours & HuGS~\cite{moreau2024human} & 29.00 & 0.9651 & 0.0635 & 27.75 & 0.9531 & 0.0694  \\
        DNA~\cite{cheng2023dna} & Ours & 27.43 & 0.9549 & 0.0779 & 25.89 & 0.9451 & 0.0864  \\ 
        DNA~\cite{cheng2023dna} & HuGS~\cite{moreau2024human} & 27.49 & 0.9558 & 0.0750 & 25.73 & 0.9427 & 0.0850  \\ \bottomrule
    \end{tabular}
    \end{adjustbox}
\end{table}

\paragraph{Setup} We select 5 sequences with challenging outfits and/or motion, consisting of 225 frames at 15 fps. While 60 cameras are available, we only use 8 for training our method, as well as 8 held-out cameras for evaluation. The ground truth SMPLx parameters of the dataset are computed with an optimization-based technique based on 3D keypoints, reported in their paper as better than EasyMocap. Measuring MPJPE over a ground truth captured by a similar technique would not be very informative, instead we propose to evaluate the accuracy of the human poses by measuring the avatar quality when an animatable body model is trained with it. We start by fitting our method on all frames of the sequences, using the same 3D keypoints as the DNA method. Then, we train avatars \textit{without} body pose optimization, but from fixed pre-computed parameters, comparing DNA parameters with ours. We use the 180 first frames to train, and keep the last 45 for novel pose evaluation. We compare our Gaussian-based avatar to Human Gaussian Splatting (HuGS)~\cite{moreau2024human}. The main difference between both models is that HuGS primitives are not defined on a mesh surface but rather rigged directly to the skeleton via learnable skinning weights. We evaluate the visual quality of avatars via PSNR, SSIM and LPIPS on novel views for both training and novel human poses. Novel view synthesis of radiance fields is commonly used to evaluate the accuracy of camera poses~\cite{brachmann2024acezero} and we argu that it is relevant for human poses as well.

\paragraph{Results} Quantitative results averaged over subjects are presented in Table~\ref{tab:DNA} and qualitative results in Figure~\ref{fig:quali_DNA_2}. The first line shows results of our joint training. Because it processes the entire video to estimate pose parameters, results on novel poses can not be evaluated. Next lines report results for models trained with pre-computed parameters, depending on pose estimation method and avatar model. First, we observe that using our pose parameters results in notably better rendering, improving novel view synthesis by 1.5dB PSNR on training poses and  2dB PSNR on novel poses, confirming the motion capture quality of our model. Qualitatively, avatars are less blurry and present better details, especially for hands. Regarding avatars, our body model performs slightly worse than HuGS on training poses of DNA, but better on novel poses. The reason is that HuGS presents more expressivity to deform the model, at the cost of worse generalization to novel poses, while our avatar is designed to rely on the pose signal only to explain details, which we observe to be important for robust pose estimation. Consequently, it performs better than HuGS using our improved parameters. Finally, our avatar exhibits better quality in the joint optimization setup, confirming the synergistic effect of solving pose estimation with avatar rendering.

\subsection{Ablation Studies}
\label{sec:ablations}

\paragraph{Number of cameras} We evaluate HPE accuracy on all sequences of MOYO dataset depending on the number of cameras used. By default, the 8 available cameras surrounding the subject were used, and we compare it to setups with 2, 4 and 6 cameras. We take care of recomputing the 3D keypoints triangulation (used in the pre-training step) using only available images for each scenario.

\begin{table}[h]
\centering
\caption{Pose Estimation accuracy against number of cameras.}
\label{tab:ablation_cams}
\begin{adjustbox}{width=\columnwidth}
\begin{tabular}{cllllll}
\toprule
 & \multicolumn{3}{c}{Body Pose Estimation} & \multicolumn{3}{c}{Hands Pose Estimation} \\
 \cmidrule(r){2-4}  \cmidrule(r){5-7}  
cams & W-MPJPE & MPJPE & PA-MPJPE & W-MPJPE & MPJPE & PA-MPJPE \\
 \cmidrule(r){1-4}  \cmidrule(r){5-7}  
8 & 32.61 & 42.23 & 24.64 & 17.90 & 14.15 & 8.16 \\
6 & 33.13 & 43.15 & 25.01 & 18.11 & 14.03 & 8.20 \\
4 & 32.84 & 43.14 & 25.22 & 18.00 & 14.32 & 8.41 \\
2 & 35.35 & 45.96 & 27.09 & 25.37 & 15.11 & 8.09 \\
\bottomrule
\end{tabular}
\end{adjustbox}
\end{table}

Results are given in Table~\ref{tab:ablation_cams}. As expected, using less cameras results in lower pose estimation accuracy. However, the decrease is suprisingly very small: a 4 cameras setup has only 0.2mm larger W-MPJPE compared to 8 cameras. Even with 2 cameras only, our method is significantly more accurate than keypoint-based methods with 8 cameras. This very promising result show the flexibility of our method to sparse camera setups.

\paragraph{Motion parametrization} To confirm the benefit of our motion MLPs, we compare it against the discrete implementation where pose parameters of each frame are given directly to the optimizer. While simpler, this formulation exhibits poor temporal consistency, because parameters of each time step are independent of the others. Thus, we also evaluate this solution with a regularization loss that encourages joints stability between consecutive timesteps. In addition to that, we try to use a human body prior, VPoser~\cite{SMPL-X:2019} instead of predicting joints rotations directly.

\begin{table}[h]
\caption{Impact of motion parametrization on pose accuracy.}
\label{tab:ablation_motion}
\begin{adjustbox}{width=\columnwidth}
\begin{tabular}{ccccccccc}
\toprule
 & & & \multicolumn{3}{c}{Body Pose Estimation}& \multicolumn{3}{c}{Hands Pose Estimation}\\
 \cmidrule(r){4-6}  \cmidrule(r){7-9}  
\textbf{MLP} & \textbf{Vposer} & \multicolumn{1}{c}{\textbf{Smoothness}} & W\_MPJPE & MPJPE & PA-MPJPE & W\_MPJPE & MPJPE & PA-MPJPE \\
\cmidrule(r){1-3} \cmidrule(r){4-6}  \cmidrule(r){7-9}  
Yes & No & No & 34.4 & 46.62 & 23.95 & 13.03 & 9.24 & 6.19 \\
Yes & Yes & No & 39.23 & 58.77 & 30.45 & 13.31 & 11.89 & 6.36 \\
No & No & No & 38.14 & 53.84 & 28.86 & 23.06 & 19.42 & 13.06 \\
No & Yes & No & 44.53 & 53.81 & 32.53 & 15.48 & 14.71 & 7.63 \\
No & No & Yes & 39.27 & 50.02 & 29.03 & 25.26 & 24.5 & 13.09 \\
No & Yes & Yes & 40.41 & 49.32 & 28.14 & 18.52 & 16.3 & 7.85 \\
\bottomrule
\end{tabular}
\end{adjustbox}
\end{table}

In Table \ref{tab:ablation_motion} we show that our motions MLPs outperform the discrete implementation with a large margin, with and without smoothness loss. Using VPoser degrades the accuracy because, despite its robustness, it is not expressive enough to fit precisely challenging yoga poses.

\paragraph{Gaussians parametrization} We compare the use of our Gaussians MLP to the explicit optimization of Gaussians parameters, similar to the original 3DGS implementation~\cite{kerbl3Dgaussians}. We compare the visual quality of avatars via NVS of training poses on DNA rendering. Results are shown in Table~\ref{tab:ablation_gaussians_MLP}. The Gaussians MLP improves metrics significantly thanks to a better spatial smoothness.

\begin{table}[h]
\caption{NVS quality with and without the Gaussians MLP.}
\label{tab:ablation_gaussians_MLP}
\centering
\footnotesize
\begin{tabular}{lccc}
\toprule
& \textbf{PSNR} & \textbf{SSIM} & \textbf{LPIPS} \\
\cmidrule(r){1-4}     
\textbf{MLP} & 29.31 & 0.9660 & 0.0655 \\
\textbf{Explicit} & 28.86 & 0.9624 & 0.0707 \\
\bottomrule
\end{tabular}
\end{table}
\section{Limitations and Future Work}
\label{sec:discussion}

One current limitation of our method is the processing time. The implementation could be further optimized and the solution can already be deployed for offline scenarios like professional motion capture and datasets annotation. One interesting direction would be to extend the idea to a single camera, following the current trend of Gaussian avatars from monocular videos~\cite{li2024gaussianbody}. Long dresses and hair are also challenging for our avatar representation because the geometry is based on a mesh. Fitting very loose clothing properly would require clothes template meshes~\cite{PhysAvatar24}. However, even for such scenarios, \textit{Better Together} estimate correct body poses. Finally, we focused on whole body motion but we believe that the proposed idea could work to track facial expression, and leave this as future work.

\paragraph{Potential Societal Impact} Photorealistic avatars have potentially dangerous applications if used maliciously to generate deepfakes videos of people without consent. Our algorithm is not designed for this purpose and we expect the community to act responsibly to prevent such use cases.
\section{Conclusion}
\label{sec:conclusion}

In this paper, we proposed a novel paradigm for markerless multi-view motion capture: instead of relying on keypoints detection and triangulation, we showed through experiments that a more precise pose estimation was possible by reconstruction of images, through the rendering of an animatable avatar. From the intuition to an effective implementation, we had to introduce several key contributions such as time-dependent neural networks to fit the sequence of human poses and a novel 3D human model with Gaussians defined on a mesh. This formulation is not only beneficial for motion capture but also for the quality of photorealistic avatars, confirming our initial hypothesis that motion capture and avatar creation are solved \textit{Better Together}.

\clearpage
{
    \small
    \bibliographystyle{ieeenat_fullname}
    \bibliography{main}

\begin{thebibliography}{81}
\providecommand{\natexlab}[1]{#1}
\providecommand{\url}[1]{\texttt{#1}}
\expandafter\ifx\csname urlstyle\endcsname\relax
  \providecommand{\doi}[1]{doi: #1}\else
  \providecommand{\doi}{doi: \begingroup \urlstyle{rm}\Url}\fi

\bibitem[eas(2021)]{easymocap}
Easymocap - make human motion capture easier.
\newblock Github, 2021.

\bibitem[Bagautdinov et~al.(2021)Bagautdinov, Wu, Simon, Prada, Shiratori, Wei, Xu, Sheikh, and Saragih]{bagautdinov2021driving}
Timur Bagautdinov, Chenglei Wu, Tomas Simon, Fabian Prada, Takaaki Shiratori, Shih-En Wei, Weipeng Xu, Yaser Sheikh, and Jason Saragih.
\newblock Driving-signal aware full-body avatars.
\newblock \emph{ACM Transactions on Graphics (TOG)}, 40\penalty0 (4):\penalty0 1--17, 2021.

\bibitem[Barron et~al.(2022)Barron, Mildenhall, Verbin, Srinivasan, and Hedman]{barron2022mipnerf360}
Jonathan~T. Barron, Ben Mildenhall, Dor Verbin, Pratul~P. Srinivasan, and Peter Hedman.
\newblock Mip-nerf 360: Unbounded anti-aliased neural radiance fields.
\newblock \emph{CVPR}, 2022.

\bibitem[Barron et~al.(2023)Barron, Mildenhall, Verbin, Srinivasan, and Hedman]{barron2023zipnerf}
Jonathan~T. Barron, Ben Mildenhall, Dor Verbin, Pratul~P. Srinivasan, and Peter Hedman.
\newblock Zip-nerf: Anti-aliased grid-based neural radiance fields.
\newblock \emph{ICCV}, 2023.

\bibitem[Bian et~al.(2023)Bian, Wang, Li, Bian, and Prisacariu]{bian2023nope}
Wenjing Bian, Zirui Wang, Kejie Li, Jia-Wang Bian, and Victor~Adrian Prisacariu.
\newblock Nope-nerf: Optimising neural radiance field with no pose prior.
\newblock In \emph{CVPR}, pages 4160--4169, 2023.

\bibitem[Brachmann et~al.(2024)Brachmann, Wynn, Chen, Cavallari, Monszpart, Turmukhambetov, and Prisacariu]{brachmann2024acezero}
Eric Brachmann, Jamie Wynn, Shuai Chen, Tommaso Cavallari, {\'{A}}ron Monszpart, Daniyar Turmukhambetov, and Victor~Adrian Prisacariu.
\newblock Scene coordinate reconstruction: Posing of image collections via incremental learning of a relocalizer.
\newblock In \emph{ECCV}, 2024.

\bibitem[Cai et~al.(2024)Cai, Heikkil\"a, and Rahtu]{cai_2024_GSPose}
Dingding Cai, Janne Heikkil\"a, and Esa Rahtu.
\newblock Gs-pose: Generalizable segmentation-based 6d object pose estimation with 3d gaussian splatting.
\newblock \emph{arXiv preprint arXiv:2403.10683v2}, 2024.

\bibitem[{Cao} et~al.(2019){Cao}, {Hidalgo Martinez}, {Simon}, {Wei}, and {Sheikh}]{openpose}
Z. {Cao}, G. {Hidalgo Martinez}, T. {Simon}, S. {Wei}, and Y.~A. {Sheikh}.
\newblock Openpose: Realtime multi-person 2d pose estimation using part affinity fields.
\newblock \emph{IEEE Transactions on Pattern Analysis and Machine Intelligence}, 2019.

\bibitem[Chen et~al.(2024)Chen, Zheng, Li, Xu, and Liu]{chen2024meshavatar}
Yushuo Chen, Zerong Zheng, Zhe Li, Chao Xu, and Yebin Liu.
\newblock Meshavatar: Learning high-quality triangular human avatars from multi-view videos.
\newblock \emph{arXiv preprint arXiv:2407.08414}, 2024.

\bibitem[Cheng et~al.(2023)Cheng, Chen, Fan, Yin, Chen, Cai, Wang, Gao, Yu, Lin, et~al.]{cheng2023dna}
Wei Cheng, Ruixiang Chen, Siming Fan, Wanqi Yin, Keyu Chen, Zhongang Cai, Jingbo Wang, Yang Gao, Zhengming Yu, Zhengyu Lin, et~al.
\newblock Dna-rendering: A diverse neural actor repository for high-fidelity human-centric rendering.
\newblock In \emph{ICCV}, pages 19982--19993, 2023.

\bibitem[Choudhury et~al.(2023)Choudhury, Kitani, and Jeni]{choudhury2023tempo}
Rohan Choudhury, Kris~M. Kitani, and Laszlo~A. Jeni.
\newblock Tempo: Efficient multi-view pose estimation, tracking, and forecasting.
\newblock In \emph{ICCV}, 2023.

\bibitem[Desmarais et~al.(2021)Desmarais, Mottet, Slangen, and Montesinos]{desmarais2021review}
Yann Desmarais, Denis Mottet, Pierre Slangen, and Philippe Montesinos.
\newblock A review of 3d human pose estimation algorithms for markerless motion capture.
\newblock \emph{Computer Vision and Image Understanding}, 212:\penalty0 103275, 2021.

\bibitem[Dhamo et~al.(2024)Dhamo, Nie, Moreau, Song, Shaw, Zhou, and P{\'e}rez-Pellitero]{dhamo2023headgas}
Helisa Dhamo, Yinyu Nie, Arthur Moreau, Jifei Song, Richard Shaw, Yiren Zhou, and Eduardo P{\'e}rez-Pellitero.
\newblock Headgas: Real-time animatable head avatars via 3d gaussian splatting.
\newblock \emph{ECCV}, 2024.

\bibitem[Dong et~al.(2019)Dong, Jiang, Huang, Bao, and Zhou]{dong2019fast}
Junting Dong, Wen Jiang, Qixing Huang, Hujun Bao, and Xiaowei Zhou.
\newblock Fast and robust multi-person 3d pose estimation from multiple views.
\newblock In \emph{CVPR}, pages 7792--7801, 2019.

\bibitem[Dwivedi et~al.(2024{\natexlab{a}})Dwivedi, Schmid, Yi, Black, and Tzionas]{dwivedi2024poco}
Sai~Kumar Dwivedi, Cordelia Schmid, Hongwei Yi, Michael~J Black, and Dimitrios Tzionas.
\newblock Poco: 3d pose and shape estimation with confidence.
\newblock In \emph{3DV}, pages 85--95, 2024{\natexlab{a}}.

\bibitem[Dwivedi et~al.(2024{\natexlab{b}})Dwivedi, Sun, Patel, Feng, and Black]{Dwivedi_2024_CVPR}
Sai~Kumar Dwivedi, Yu Sun, Priyanka Patel, Yao Feng, and Michael~J. Black.
\newblock Tokenhmr: Advancing human mesh recovery with a tokenized pose representation.
\newblock In \emph{CVPR}, pages 1323--1333, 2024{\natexlab{b}}.

\bibitem[Fu et~al.(2024)Fu, Liu, Kulkarni, Kautz, Efros, and Wang]{Fu_2024_CVPR}
Yang Fu, Sifei Liu, Amey Kulkarni, Jan Kautz, Alexei~A. Efros, and Xiaolong Wang.
\newblock Colmap-free 3d gaussian splatting.
\newblock In \emph{CVPR}, pages 20796--20805, 2024.

\bibitem[Goel et~al.(2023)Goel, Pavlakos, Rajasegaran, Kanazawa*, and Malik*]{goel2023humans}
Shubham Goel, Georgios Pavlakos, Jathushan Rajasegaran, Angjoo Kanazawa*, and Jitendra Malik*.
\newblock Humans in 4{D}: Reconstructing and tracking humans with transformers.
\newblock In \emph{ICCV}, 2023.

\bibitem[Gu{\'e}don and Lepetit(2024)]{guedon2024frosting}
Antoine Gu{\'e}don and Vincent Lepetit.
\newblock Gaussian frosting: Editable complex radiance fields with real-time rendering.
\newblock \emph{ECCV}, 2024.

\bibitem[Guo et~al.(2024)Guo, Kumar, Zhao, Wang, Huang, and Ren]{Guo2024ECCV}
Yuliang Guo, Abhinav Kumar, Cheng Zhao, Ruoyu Wang, Xinyu Huang, and Liu Ren.
\newblock Sup-nerf: A streamlined unification of pose estimation and nerf for monocular 3d object reconstruction.
\newblock \emph{ECCV}, 2024.

\bibitem[He et~al.(2022)He, Saito, Zachary, Rushmeier, and Zhou]{he2022nemf}
Chengan He, Jun Saito, James Zachary, Holly Rushmeier, and Yi Zhou.
\newblock Nemf: Neural motion fields for kinematic animation.
\newblock \emph{Advances in Neural Information Processing Systems}, 35:\penalty0 4244--4256, 2022.

\bibitem[Henderson et~al.(2020)Henderson, Tsiminaki, and Lampert]{henderson2020leveraging}
Paul Henderson, Vagia Tsiminaki, and Christoph~H Lampert.
\newblock Leveraging 2d data to learn textured 3d mesh generation.
\newblock In \emph{CVPR}, pages 7498--7507, 2020.

\bibitem[Herau et~al.(2023)Herau, Piasco, Bennehar, Rold{\~a}o, Tsishkou, Migniot, Vasseur, and Demonceaux]{herau2023moisst}
Quentin Herau, Nathan Piasco, Moussab Bennehar, Luis Rold{\~a}o, Dzmitry Tsishkou, Cyrille Migniot, Pascal Vasseur, and C{\'e}dric Demonceaux.
\newblock Moisst: Multimodal optimization of implicit scene for spatiotemporal calibration.
\newblock In \emph{2023 IEEE/RSJ International Conference on Intelligent Robots and Systems (IROS)}, pages 1810--1817, 2023.

\bibitem[Herau et~al.(2024{\natexlab{a}})Herau, Bennehar, Moreau, Piasco, Rold{\~a}o, Tsishkou, Migniot, Vasseur, and Demonceaux]{herau20243dgs}
Quentin Herau, Moussab Bennehar, Arthur Moreau, Nathan Piasco, Luis Rold{\~a}o, Dzmitry Tsishkou, Cyrille Migniot, Pascal Vasseur, and C{\'e}dric Demonceaux.
\newblock 3dgs-calib: 3d gaussian splatting for multimodal spatiotemporal calibration.
\newblock In \emph{2024 IEEE/RSJ International Conference on Intelligent Robots and Systems (IROS)}, pages 8315--8321. IEEE, 2024{\natexlab{a}}.

\bibitem[Herau et~al.(2024{\natexlab{b}})Herau, Piasco, Bennehar, Roldao, Tsishkou, Migniot, Vasseur, and Demonceaux]{herau2024soac}
Quentin Herau, Nathan Piasco, Moussab Bennehar, Luis Roldao, Dzmitry Tsishkou, Cyrille Migniot, Pascal Vasseur, and C{\'e}dric Demonceaux.
\newblock Soac: Spatio-temporal overlap-aware multi-sensor calibration using neural radiance fields.
\newblock In \emph{CVPR}, pages 15131--15140, 2024{\natexlab{b}}.

\bibitem[Hu et~al.(2024)Hu, Zhang, Zhang, Zhou, Liu, Zhang, and Nie]{hu2024gaussianavatar}
Liangxiao Hu, Hongwen Zhang, Yuxiang Zhang, Boyao Zhou, Boning Liu, Shengping Zhang, and Liqiang Nie.
\newblock Gaussianavatar: Towards realistic human avatar modeling from a single video via animatable 3d gaussians.
\newblock In \emph{CVPR}, pages 634--644, 2024.

\bibitem[Huang et~al.(2017)Huang, Bogo, Lassner, Kanazawa, Gehler, Romero, Akhter, and Black]{MuVS:3DV:2017}
Yinghao Huang, Federica Bogo, Christoph Lassner, Angjoo Kanazawa, Peter~V. Gehler, Javier Romero, Ijaz Akhter, and Michael~J. Black.
\newblock Towards accurate marker-less human shape and pose estimation over time.
\newblock In \emph{3DV}, pages 421--430, 2017.

\bibitem[Iskakov et~al.(2019)Iskakov, Burkov, Lempitsky, and Malkov]{iskakov2019learnable}
Karim Iskakov, Egor Burkov, Victor Lempitsky, and Yury Malkov.
\newblock Learnable triangulation of human pose.
\newblock In \emph{ICCV}, 2019.

\bibitem[Jiang et~al.(2024)Jiang, Shen, Wang, Su, Hong, Zhang, Yu, and Xu]{jiang2024hifi4g}
Yuheng Jiang, Zhehao Shen, Penghao Wang, Zhuo Su, Yu Hong, Yingliang Zhang, Jingyi Yu, and Lan Xu.
\newblock Hifi4g: High-fidelity human performance rendering via compact gaussian splatting.
\newblock In \emph{CVPR}, pages 19734--19745, 2024.

\bibitem[Kanazawa et~al.(2018)Kanazawa, Black, Jacobs, and Malik]{hmrKanazawa17}
Angjoo Kanazawa, Michael~J. Black, David~W. Jacobs, and Jitendra Malik.
\newblock End-to-end recovery of human shape and pose.
\newblock In \emph{CVPR}, 2018.

\bibitem[Kerbl et~al.(2023)Kerbl, Kopanas, Leimk{\"u}hler, and Drettakis]{kerbl3Dgaussians}
Bernhard Kerbl, Georgios Kopanas, Thomas Leimk{\"u}hler, and George Drettakis.
\newblock 3d gaussian splatting for real-time radiance field rendering.
\newblock \emph{ACM Transactions on Graphics}, 42\penalty0 (4), 2023.

\bibitem[Khirodkar et~al.(2025)Khirodkar, Bagautdinov, Martinez, Zhaoen, James, Selednik, Anderson, and Saito]{khirodkar2025sapiens}
Rawal Khirodkar, Timur Bagautdinov, Julieta Martinez, Su Zhaoen, Austin James, Peter Selednik, Stuart Anderson, and Shunsuke Saito.
\newblock Sapiens: Foundation for human vision models.
\newblock In \emph{European Conference on Computer Vision}, pages 206--228, 2025.

\bibitem[Kocabas et~al.(2024)Kocabas, Yuan, Molchanov, Guo, Black, Hilliges, Kautz, and Iqbal]{kocabas2024pace}
Muhammed Kocabas, Ye Yuan, Pavlo Molchanov, Yunrong Guo, Michael~J Black, Otmar Hilliges, Jan Kautz, and Umar Iqbal.
\newblock Pace: Human and camera motion estimation from in-the-wild videos.
\newblock In \emph{3DV}, pages 397--408, 2024.

\bibitem[Kolotouros et~al.(2021)Kolotouros, Pavlakos, Jayaraman, and Daniilidis]{kolotouros2019spin}
Nikos Kolotouros, Georgios Pavlakos, Dinesh Jayaraman, and Kostas Daniilidis.
\newblock Probabilistic modeling for human mesh recovery.
\newblock In \emph{ICCV}, 2021.

\bibitem[Li et~al.(2024{\natexlab{a}})Li, Yao, Xie, Chen, and Jiang]{li2024gaussianbody}
Mengtian Li, Shengxiang Yao, Zhifeng Xie, Keyu Chen, and Yu-Gang Jiang.
\newblock Gaussianbody: Clothed human reconstruction via 3d gaussian splatting.
\newblock \emph{arXiv preprint arXiv:2401.09720}, 2024{\natexlab{a}}.

\bibitem[Li et~al.(2022)Li, Tanke, Vo, Zollhofer, Gall, Kanazawa, and Lassner]{li2022tava}
Ruilong Li, Julian Tanke, Minh Vo, Michael Zollhofer, Jurgen Gall, Angjoo Kanazawa, and Christoph Lassner.
\newblock Tava: Template-free animatable volumetric actors.
\newblock In \emph{ECCV}, 2022.

\bibitem[Li et~al.(2021)Li, Niklaus, Snavely, and Wang]{li2020neural}
Zhengqi Li, Simon Niklaus, Noah Snavely, and Oliver Wang.
\newblock Neural scene flow fields for space-time view synthesis of dynamic scenes.
\newblock In \emph{CVPR}, 2021.

\bibitem[Li et~al.(2024{\natexlab{b}})Li, Zheng, Wang, and Liu]{li2024animatable}
Zhe Li, Zerong Zheng, Lizhen Wang, and Yebin Liu.
\newblock Animatable gaussians: Learning pose-dependent gaussian maps for high-fidelity human avatar modeling.
\newblock In \emph{CVPR}, pages 19711--19722, 2024{\natexlab{b}}.

\bibitem[Liao et~al.(2024)Liao, Zhu, Wang, Hu, and Waslander]{liao2024multiple}
Ziwei Liao, Jialiang Zhu, Chunyu Wang, Han Hu, and Steven~L Waslander.
\newblock Multiple view geometry transformers for 3d human pose estimation.
\newblock In \emph{CVPR}, pages 708--717, 2024.

\bibitem[Lin et~al.(2021)Lin, Wang, and Liu]{lin2021mesh}
Kevin Lin, Lijuan Wang, and Zicheng Liu.
\newblock Mesh graphormer.
\newblock In \emph{ICCV}, pages 12939--12948, 2021.

\bibitem[Loper et~al.(2015)Loper, Mahmood, Romero, Pons-Moll, and Black]{SMPL:2015}
Matthew Loper, Naureen Mahmood, Javier Romero, Gerard Pons-Moll, and Michael~J. Black.
\newblock {SMPL}: A skinned multi-person linear model.
\newblock \emph{SIGGRAPH Asia}, 34\penalty0 (6):\penalty0 248:1--248:16, 2015.

\bibitem[Lu et~al.(2024)Lu, Dong, Song, and Hilliges]{lu2024avatarpose}
Feichi Lu, Zijian Dong, Jie Song, and Otmar Hilliges.
\newblock Avatarpose: Avatar-guided 3d pose estimation of close human interaction from sparse multi-view videos.
\newblock In \emph{ECCV}, 2024.

\bibitem[Matsuki et~al.(2024)Matsuki, Murai, Kelly, and Davison]{Matsuki:Murai:etal:CVPR2024}
Hidenobu Matsuki, Riku Murai, Paul H.~J. Kelly, and Andrew~J. Davison.
\newblock {G}aussian {S}platting {SLAM}.
\newblock \emph{CVPR}, 2024.

\bibitem[Menolotto et~al.(2020)Menolotto, Komaris, Tedesco, O’Flynn, and Walsh]{menolotto2020motion}
Matteo Menolotto, Dimitrios-Sokratis Komaris, Salvatore Tedesco, Brendan O’Flynn, and Michael Walsh.
\newblock Motion capture technology in industrial applications: A systematic review.
\newblock \emph{Sensors}, 20\penalty0 (19):\penalty0 5687, 2020.

\bibitem[Mildenhall et~al.(2020)Mildenhall, Srinivasan, Tancik, Barron, Ramamoorthi, and Ng]{nerf}
Ben Mildenhall, Pratul~P. Srinivasan, Matthew Tancik, Jonathan~T. Barron, Ravi Ramamoorthi, and Ren Ng.
\newblock Nerf: Representing scenes as neural radiance fields for view synthesis.
\newblock In \emph{ECCV}, 2020.

\bibitem[Moreau et~al.(2022)Moreau, Piasco, Tsishkou, Stanciulescu, and de~La~Fortelle]{lens}
Arthur Moreau, Nathan Piasco, Dzmitry Tsishkou, Bogdan Stanciulescu, and Arnaud de La~Fortelle.
\newblock Lens: Localization enhanced by nerf synthesis.
\newblock In \emph{Proceedings of the 5th Conference on Robot Learning}, pages 1347--1356. PMLR, 2022.

\bibitem[Moreau et~al.(2023)Moreau, Piasco, Bennehar, Tsishkou, Stanciulescu, and de~La~Fortelle]{moreau2023crossfire}
Arthur Moreau, Nathan Piasco, Moussab Bennehar, Dzmitry Tsishkou, Bogdan Stanciulescu, and Arnaud de La~Fortelle.
\newblock Crossfire: Camera relocalization on self-supervised features from an implicit representation.
\newblock In \emph{ICCV}, pages 252--262, 2023.

\bibitem[Moreau et~al.(2024)Moreau, Song, Dhamo, Shaw, Zhou, and P{\'e}rez-Pellitero]{moreau2024human}
Arthur Moreau, Jifei Song, Helisa Dhamo, Richard Shaw, Yiren Zhou, and Eduardo P{\'e}rez-Pellitero.
\newblock Human gaussian splatting: Real-time rendering of animatable avatars.
\newblock In \emph{CVPR}, pages 788--798, 2024.

\bibitem[M\"uller et~al.(2022)M\"uller, Evans, Schied, and Keller]{mueller2022instant}
Thomas M\"uller, Alex Evans, Christoph Schied, and Alexander Keller.
\newblock Instant neural graphics primitives with a multiresolution hash encoding.
\newblock \emph{ACM Trans. Graph.}, 41\penalty0 (4):\penalty0 102:1--102:15, 2022.

\bibitem[Nogueira et~al.(2024)Nogueira, Oliveira, and Teixeira]{nogueira2024markerless}
Ana Filipa~Rodrigues Nogueira, H{\'e}lder~P Oliveira, and Lu{\'\i}s~F Teixeira.
\newblock Markerless multi-view 3d human pose estimation: a survey.
\newblock \emph{arXiv preprint arXiv:2407.03817}, 2024.

\bibitem[Pang et~al.(2024)Pang, Zhu, Kortylewski, Theobalt, and Habermann]{pang2024ash}
Haokai Pang, Heming Zhu, Adam Kortylewski, Christian Theobalt, and Marc Habermann.
\newblock Ash: Animatable gaussian splats for efficient and photoreal human rendering.
\newblock In \emph{CVPR}, pages 1165--1175, 2024.

\bibitem[Park et~al.(2021)Park, Sinha, Barron, Bouaziz, Goldman, Seitz, and Martin-Brualla]{park2021nerfies}
Keunhong Park, Utkarsh Sinha, Jonathan~T Barron, Sofien Bouaziz, Dan~B Goldman, Steven~M Seitz, and Ricardo Martin-Brualla.
\newblock Nerfies: Deformable neural radiance fields.
\newblock In \emph{ICCV}, pages 5865--5874, 2021.

\bibitem[Paszke et~al.(2019)Paszke, Gross, Massa, Lerer, Bradbury, Chanan, Killeen, Lin, Gimelshein, Antiga, et~al.]{paszke2019pytorch}
Adam Paszke, Sam Gross, Francisco Massa, Adam Lerer, James Bradbury, Gregory Chanan, Trevor Killeen, Zeming Lin, Natalia Gimelshein, Luca Antiga, et~al.
\newblock Pytorch: An imperative style, high-performance deep learning library.
\newblock \emph{Advances in neural information processing systems}, 32, 2019.

\bibitem[Pavlakos et~al.(2019)Pavlakos, Choutas, Ghorbani, Bolkart, Osman, Tzionas, and Black]{SMPL-X:2019}
Georgios Pavlakos, Vasileios Choutas, Nima Ghorbani, Timo Bolkart, Ahmed A.~A. Osman, Dimitrios Tzionas, and Michael~J. Black.
\newblock Expressive body capture: {3D} hands, face, and body from a single image.
\newblock In \emph{CVPR}, pages 10975--10985, 2019.

\bibitem[Peng et~al.(2021{\natexlab{a}})Peng, Dong, Wang, Zhang, Shuai, Zhou, and Bao]{peng2021animatable}
Sida Peng, Junting Dong, Qianqian Wang, Shangzhan Zhang, Qing Shuai, Xiaowei Zhou, and Hujun Bao.
\newblock Animatable neural radiance fields for modeling dynamic human bodies.
\newblock In \emph{ICCV}, 2021{\natexlab{a}}.

\bibitem[Peng et~al.(2021{\natexlab{b}})Peng, Zhang, Xu, Wang, Shuai, Bao, and Zhou]{peng2021neural}
Sida Peng, Yuanqing Zhang, Yinghao Xu, Qianqian Wang, Qing Shuai, Hujun Bao, and Xiaowei Zhou.
\newblock Neural body: Implicit neural representations with structured latent codes for novel view synthesis of dynamic humans.
\newblock In \emph{CVPR}, 2021{\natexlab{b}}.

\bibitem[Qian et~al.(2024{\natexlab{a}})Qian, Kirschstein, Schoneveld, Davoli, Giebenhain, and Nie{\ss}ner]{qian2024gaussianavatars}
Shenhan Qian, Tobias Kirschstein, Liam Schoneveld, Davide Davoli, Simon Giebenhain, and Matthias Nie{\ss}ner.
\newblock Gaussianavatars: Photorealistic head avatars with rigged 3d gaussians.
\newblock In \emph{CVPR}, pages 20299--20309, 2024{\natexlab{a}}.

\bibitem[Qian et~al.(2024{\natexlab{b}})Qian, Wang, Mihajlovic, Geiger, and Tang]{qian20243dgs}
Zhiyin Qian, Shaofei Wang, Marko Mihajlovic, Andreas Geiger, and Siyu Tang.
\newblock 3dgs-avatar: Animatable avatars via deformable 3d gaussian splatting.
\newblock In \emph{CVPR}, pages 5020--5030, 2024{\natexlab{b}}.

\bibitem[Ravi et~al.(2024)Ravi, Gabeur, Hu, Hu, Ryali, Ma, Khedr, R{\"a}dle, Rolland, Gustafson, Mintun, Pan, Alwala, Carion, Wu, Girshick, Doll{\'a}r, and Feichtenhofer]{ravi2024sam2}
Nikhila Ravi, Valentin Gabeur, Yuan-Ting Hu, Ronghang Hu, Chaitanya Ryali, Tengyu Ma, Haitham Khedr, Roman R{\"a}dle, Chloe Rolland, Laura Gustafson, Eric Mintun, Junting Pan, Kalyan~Vasudev Alwala, Nicolas Carion, Chao-Yuan Wu, Ross Girshick, Piotr Doll{\'a}r, and Christoph Feichtenhofer.
\newblock Sam 2: Segment anything in images and videos.
\newblock \emph{arXiv preprint arXiv:2408.00714}, 2024.

\bibitem[Remelli et~al.(2020)Remelli, Han, Honari, Fua, and Wang]{remelli2020lightweight}
Edoardo Remelli, Shangchen Han, Sina Honari, Pascal Fua, and Robert Wang.
\newblock Lightweight multi-view 3d pose estimation through camera-disentangled representation.
\newblock In \emph{CVPR}, pages 6040--6049, 2020.

\bibitem[Romero et~al.(2017)Romero, Tzionas, and Black]{MANO:SIGGRAPHASIA:2017}
Javier Romero, Dimitrios Tzionas, and Michael~J. Black.
\newblock Embodied hands: Modeling and capturing hands and bodies together.
\newblock \emph{ACM Transactions on Graphics, (Proc. SIGGRAPH Asia)}, 36\penalty0 (6), 2017.

\bibitem[Rosinol et~al.(2023)Rosinol, Leonard, and Carlone]{rosinol2023nerf}
Antoni Rosinol, John~J Leonard, and Luca Carlone.
\newblock Nerf-slam: Real-time dense monocular slam with neural radiance fields.
\newblock In \emph{2023 IEEE/RSJ International Conference on Intelligent Robots and Systems (IROS)}, pages 3437--3444, 2023.

\bibitem[Saito et~al.(2021)Saito, Yang, Ma, and Black]{Saito:CVPR:2021}
Shunsuke Saito, Jinlong Yang, Qianli Ma, and Michael~J. Black.
\newblock {SCANimate}: Weakly supervised learning of skinned clothed avatar networks.
\newblock In \emph{CVPR}, 2021.

\bibitem[Seidenschwarz et~al.(2024)Seidenschwarz, Zhou, Duisterhof, Ramanan, and Leal-Taix{\'e}]{seidenschwarz2024dynomo}
Jenny Seidenschwarz, Qunjie Zhou, Bardienus Duisterhof, Deva Ramanan, and Laura Leal-Taix{\'e}.
\newblock Dynomo: Online point tracking by dynamic online monocular gaussian reconstruction.
\newblock \emph{arXiv preprint arXiv:2409.02104}, 2024.

\bibitem[Shaw et~al.(2024)Shaw, Nazarczuk, Song, Moreau, Catley-Chandar, Dhamo, and P{\'e}rez-Pellitero]{shawswings}
Richard Shaw, Michal Nazarczuk, Jifei Song, Arthur Moreau, Sibi Catley-Chandar, Helisa Dhamo, and Eduardo P{\'e}rez-Pellitero.
\newblock Swings: Sliding windows for dynamic 3d gaussian splatting.
\newblock \emph{ECCV}, 2024.

\bibitem[Stathopoulos et~al.(2024)Stathopoulos, Han, and Metaxas]{stathopoulos2024score}
Anastasis Stathopoulos, Ligong Han, and Dimitris Metaxas.
\newblock Score-guided diffusion for 3d human recovery.
\newblock In \emph{CVPR}, 2024.

\bibitem[Stoll et~al.(2011)Stoll, Hasler, Gall, Seidel, and Theobalt]{stoll2011fast}
Carsten Stoll, Nils Hasler, Juergen Gall, Hans-Peter Seidel, and Christian Theobalt.
\newblock Fast articulated motion tracking using a sums of gaussians body model.
\newblock In \emph{ICCV}, pages 951--958, 2011.

\bibitem[Tripathi et~al.(2023)Tripathi, M{\"u}ller, Huang, Omid, Black, and Tzionas]{tripathi2023ipman}
Shashank Tripathi, Lea M{\"u}ller, Chun-Hao~P. Huang, Taheri Omid, Michael~J. Black, and Dimitrios Tzionas.
\newblock {3D} human pose estimation via intuitive physics.
\newblock In \emph{CVPR}, pages 4713--4725, 2023.

\bibitem[Tu et~al.(2020)Tu, Wang, and Zeng]{tu2020voxelpose}
Hanyue Tu, Chunyu Wang, and Wenjun Zeng.
\newblock Voxelpose: Towards multi-camera 3d human pose estimation in wild environment.
\newblock In \emph{ECCV}, pages 197--212, 2020.

\bibitem[Wang et~al.(2004)Wang, Bovik, Sheikh, and Simoncelli]{wang2004image}
Zhou Wang, Alan~C Bovik, Hamid~R Sheikh, and Eero~P Simoncelli.
\newblock Image quality assessment: from error visibility to structural similarity.
\newblock \emph{IEEE Transactions on Image Processing}, 13\penalty0 (4):\penalty0 600--612, 2004.

\bibitem[Wen et~al.(2024)Wen, Zhao, Ren, Schwing, and Wang]{wen2024gomavatar}
Jing Wen, Xiaoming Zhao, Zhongzheng Ren, Alex Schwing, and Shenlong Wang.
\newblock {GoMAvatar: Efficient Animatable Human Modeling from Monocular Video Using Gaussians-on-Mesh}.
\newblock In \emph{CVPR}, 2024.

\bibitem[Wu et~al.(2024)Wu, Yi, Fang, Xie, Zhang, Wei, Liu, Tian, and Wang]{Wu_2024_CVPR}
Guanjun Wu, Taoran Yi, Jiemin Fang, Lingxi Xie, Xiaopeng Zhang, Wei Wei, Wenyu Liu, Qi Tian, and Xinggang Wang.
\newblock 4d gaussian splatting for real-time dynamic scene rendering.
\newblock In \emph{CVPR}, pages 20310--20320, 2024.

\bibitem[Xiang et~al.(2019)Xiang, Joo, and Sheikh]{xiang2019monocular}
Donglai Xiang, Hanbyul Joo, and Yaser Sheikh.
\newblock Monocular total capture: Posing face, body, and hands in the wild.
\newblock In \emph{CVPR}, pages 10965--10974, 2019.

\bibitem[Xu et~al.(2022)Xu, Zhang, Zhang, and Tao]{xu2022vitpose}
Yufei Xu, Jing Zhang, Qiming Zhang, and Dacheng Tao.
\newblock Vi{TP}ose: Simple vision transformer baselines for human pose estimation.
\newblock In \emph{Advances in Neural Information Processing Systems}, 2022.

\bibitem[Yang et~al.(2021)Yang, Zhan, Li, Xu, Li, and Lu]{yang2021cpf}
Lixin Yang, Xinyu Zhan, Kailin Li, Wenqiang Xu, Jiefeng Li, and Cewu Lu.
\newblock {CPF}: Learning a contact potential field to model the hand-object interaction.
\newblock In \emph{ICCV}, 2021.

\bibitem[Yen-Chen et~al.(2021)Yen-Chen, Florence, Barron, Rodriguez, Isola, and Lin]{yen2020inerf}
Lin Yen-Chen, Pete Florence, Jonathan~T. Barron, Alberto Rodriguez, Phillip Isola, and Tsung-Yi Lin.
\newblock {iNeRF}: Inverting neural radiance fields for pose estimation.
\newblock In \emph{IEEE/RSJ International Conference on Intelligent Robots and Systems ({IROS})}, 2021.

\bibitem[Zhang et~al.(2021)Zhang, Cai, Yan, Feng, et~al.]{zhang2021direct}
Jianfeng Zhang, Yujun Cai, Shuicheng Yan, Jiashi Feng, et~al.
\newblock Direct multi-view multi-person 3d pose estimation.
\newblock \emph{Advances in Neural Information Processing Systems}, 34:\penalty0 13153--13164, 2021.

\bibitem[Zhang et~al.(2018)Zhang, Isola, Efros, Shechtman, and Wang]{zhang2018unreasonable}
Richard Zhang, Phillip Isola, Alexei~A Efros, Eli Shechtman, and Oliver Wang.
\newblock The unreasonable effectiveness of deep features as a perceptual metric.
\newblock In \emph{CVPR}, pages 586--595, 2018.

\bibitem[Zheng et~al.(2024)Zheng, Zhao, Yang, Yifan, Xiang, Dubost, Lagun, Beeler, Tombari, Guibas, and Wetzstein]{PhysAvatar24}
Yang Zheng, Qingqing Zhao, Guandao Yang, Wang Yifan, Donglai Xiang, Florian Dubost, Dmitry Lagun, Thabo Beeler, Federico Tombari, Leonidas Guibas, and Gordon Wetzstein.
\newblock Physavatar: Learning the physics of dressed 3d avatars from visual observations.
\newblock \emph{arxiv}, 2024.

\bibitem[Zheng et~al.(2021)Zheng, Yu, Liu, and Dai]{zheng2020pamir}
Zerong Zheng, Tao Yu, Yebin Liu, and Qionghai Dai.
\newblock Pamir: Parametric model-conditioned implicit representation for image-based human reconstruction.
\newblock \emph{IEEE Transactions on Pattern Analysis and Machine Intelligence}, pages 1--1, 2021.

\bibitem[Zheng et~al.(2022)Zheng, Huang, Yu, Zhang, Guo, and Liu]{SLRF}
Zerong Zheng, Han Huang, Tao Yu, Hongwen Zhang, Yandong Guo, and Yebin Liu.
\newblock Structured local radiance fields for human avatar modeling.
\newblock In \emph{CVPR}, 2022.

\end{thebibliography}
}

\clearpage
\setcounter{page}{1}
\maketitlesupplementary

\begin{figure}[h!]
\centering
\includegraphics[width=0.99\columnwidth]{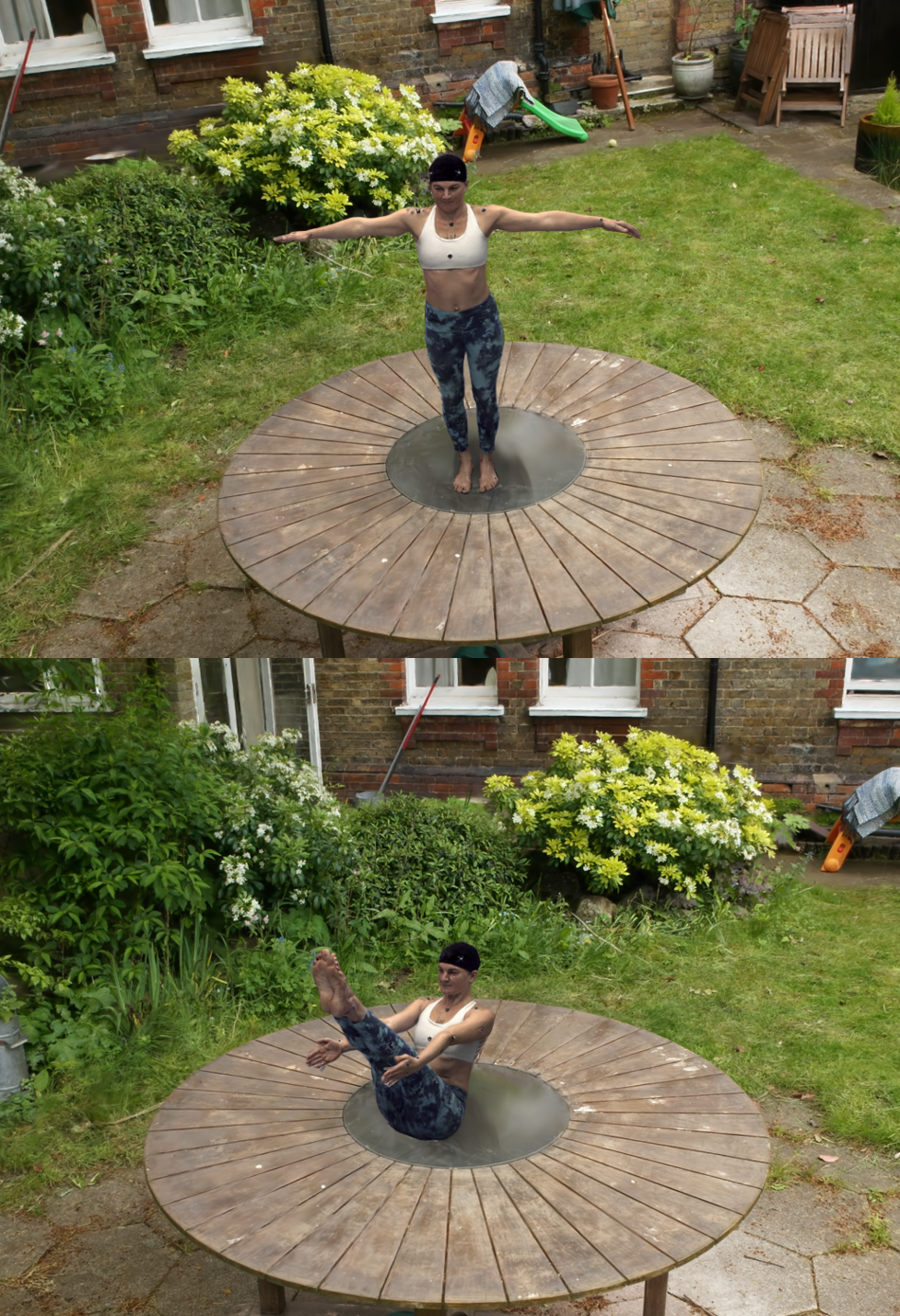}
\caption{Integration of our avatar in an external Gaussian splatting scene from Mip-NeRF360~\cite{barron2022mipnerf360}.}
\label{fig:supp_integration}
\end{figure}

\section{Supplementary video}

We invite readers to check the supplementary video that showcases our results better than just static images. The attached video contains 3 parts. First, we show a visualization of the Human Pose Estimation benchmark on the MOYO dataset. Renderings of template meshes are displayed for our method, as well as 3 baselines. We observe that for these challenging yoga poses, optimization methods are quite unstable because keypoint detectors are not precise enough (even with the latest foundation model Sapiens~\cite{khirodkar2025sapiens}). Uncommon body poses generate failures on all baselines because they are out-of-distribution of training datasets used to train regressors and keypoints detectors. Our method performs better because it optimizes the pose using dense photometric supervision.

Then, we show Novel View Synthesis of the avatar reconstructed by our method on the \textit{Boat Pose} video of the MOYO dataset. We propose a demo where the avatar is integrated in an external Gaussian splatting scene from the Mip-NeRF360~\cite{barron2022mipnerf360} dataset. Selected frames can be visualized with better quality in Figure~\ref{fig:supp_integration}. Novel views can be rendered robustly, jointly with the background, in real-time ($\approx 40$fps on a V100 GPU), and without any obvious artifact.

Finally, we show motion transfer ability, where we export the human motions learned from the DNA-Rendering dataset to the yoga subject. While the quality is slightly reduced compared to replay, we show that our method is able to extrapolate to a wide range of motion with a good quality.

\section{More details on experimental results}

\begin{table}[t]
    \centering
    \caption{Per subject NVS evaluation on DNA-Rendering.}
    \label{tab:supp_DNA_subjects}
    \subcaption*{\textbf{0034 04 (Dance)}}
    \begin{adjustbox}{width=\columnwidth}
    \begin{tabular}{llcccccr}
    \toprule
         &  & \multicolumn{3}{c}{\textit{Training Poses}} & \multicolumn{3}{c}{\textit{Novel Poses}} \\
        \cmidrule(r){3-5}  \cmidrule(lr){6-8}
        Pose & Avatar & \textbf{PSNR} & \textbf{SSIM} & \textbf{LPIPS} & \textbf{PSNR} & \textbf{SSIM} & \textbf{LPIPS}  \\ 
        \cmidrule(r){1-5}  \cmidrule(lr){6-8}
        \multicolumn{2}{c}{\textit{Better Together}} & 28.07 & 0.9584 & 0.0506 &  \multicolumn{3}{c}{-}  \\
        \midrule
        Ours & Ours & 27.85 & 0.9567 & 0.0507 & 27.20 & 0.9513 & 0.0546  \\
        Ours & HuGS~\cite{moreau2024human} & 28.01 & 0.9576 & 0.0489 & 26.85 & 0.9495 & 0.0546  \\
        DNA~\cite{cheng2023dna} & Ours & 25.85 & 0.9431 & 0.0620 & 24.58 & 0.9336 & 0.0721  \\ 
        DNA~\cite{cheng2023dna} & HuGS~\cite{moreau2024human} & 25.94 & 0.9444 & 0.0611 & 24.02 & 0.9276 & 0.0764  \\ \bottomrule
    \end{tabular}
    \end{adjustbox}
    \subcaption*{\textbf{0017 11 (Jump)}}
    \begin{adjustbox}{width=\columnwidth}
    \begin{tabular}{llcccccr}
    \toprule
         &  & \multicolumn{3}{c}{\textit{Training Poses}} & \multicolumn{3}{c}{\textit{Novel Poses}} \\
        \cmidrule(r){3-5}  \cmidrule(lr){6-8}
        Pose & Avatar & \textbf{PSNR} & \textbf{SSIM} & \textbf{LPIPS} & \textbf{PSNR} & \textbf{SSIM} & \textbf{LPIPS}  \\ 
        \cmidrule(r){1-5}  \cmidrule(lr){6-8}
        \multicolumn{2}{c}{\textit{Better Together}} & 27.32 & 0.9709 & 0.0604 &  \multicolumn{3}{c}{-}  \\
        \midrule
        Ours & Ours & 27.33 & 0.9716 & 0.0585 & 25.39 & 0.9643 & 0.0616  \\
        Ours & HuGS~\cite{moreau2024human} & 27.08 & 0.9708 & 0.0599 & 25.26 & 0.9635 & 0.0639  \\
        DNA~\cite{cheng2023dna} & Ours & 24.97 & 0.9604 & 0.0719 & 23.50 & 0.9542 & 0.0742  \\ 
        DNA~\cite{cheng2023dna} & HuGS~\cite{moreau2024human} & 25.28 & 0.9627 & 0.0698 & 23.35 & 0.9529 & 0.0764  \\ \bottomrule
    \end{tabular}
    \end{adjustbox}
    \subcaption*{\textbf{0173 08 (Fashion parade)}}
    \begin{adjustbox}{width=\columnwidth}
    \begin{tabular}{llcccccr}
    \toprule
         &  & \multicolumn{3}{c}{\textit{Training Poses}} & \multicolumn{3}{c}{\textit{Novel Poses}} \\
        \cmidrule(r){3-5}  \cmidrule(lr){6-8}
        Pose & Avatar & \textbf{PSNR} & \textbf{SSIM} & \textbf{LPIPS} & \textbf{PSNR} & \textbf{SSIM} & \textbf{LPIPS}  \\ 
        \cmidrule(r){1-5}  \cmidrule(lr){6-8}
        \multicolumn{2}{c}{\textit{Better Together}} & 31.00 & 0.9783 & 0.0465 &  \multicolumn{3}{c}{-}  \\
        \midrule
        Ours & Ours & 30.60 & 0.9773 & 0.0462 & 28.65 & 0.9694 & 0.0528  \\
        Ours & HuGS~\cite{moreau2024human} & 30.11 & 0.9769 & 0.0462 & 28.72 & 0.9696 & 0.0512  \\
        DNA~\cite{cheng2023dna} & Ours & 28.85 & 0.9703 & 0.0530 & 27.10 & 0.9600 & 0.0612  \\ 
        DNA~\cite{cheng2023dna} & HuGS~\cite{moreau2024human} & 28.74 & 0.9702 & 0.0526 & 26.80 & 0.9595 & 0.0601  \\ \bottomrule
    \end{tabular}
    \end{adjustbox}
    \subcaption*{\textbf{0235 11 (Fireman outfit)}}
    \begin{adjustbox}{width=\columnwidth}
    \begin{tabular}{llcccccr}
    \toprule
         &  & \multicolumn{3}{c}{\textit{Training Poses}} & \multicolumn{3}{c}{\textit{Novel Poses}} \\
        \cmidrule(r){3-5}  \cmidrule(lr){6-8}
        Pose & Avatar & \textbf{PSNR} & \textbf{SSIM} & \textbf{LPIPS} & \textbf{PSNR} & \textbf{SSIM} & \textbf{LPIPS}  \\ 
        \cmidrule(r){1-5}  \cmidrule(lr){6-8}
        \multicolumn{2}{c}{\textit{Better Together}} & 28.75 & 0.9637 & 0.0643 &  \multicolumn{3}{c}{-}  \\
        \midrule
        Ours & Ours & 28.67 & 0.9626 & 0.0645 & 26.33 & 0.9484 & 0.0757  \\
        Ours & HuGS~\cite{moreau2024human} & 28.43 & 0.9619 & 0.0621 & 26.37 & 0.9486 & 0.0757  \\
        DNA~\cite{cheng2023dna} & Ours & 27.41 & 0.9526 & 0.0753 & 25.10 & 0.9366 & 0.0931  \\ 
        DNA~\cite{cheng2023dna} & HuGS~\cite{moreau2024human} & 27.48 & 0.9544 & 0.0705 & 25.15 & 0.9320 & 0.0893  \\ \bottomrule
    \end{tabular}
    \end{adjustbox}

    \subcaption*{\textbf{0152 01 (Kneeling with a costume)}}
    \begin{adjustbox}{width=\columnwidth}
    
    \begin{tabular}{llcccccr}
    \toprule
         &  & \multicolumn{3}{c}{\textit{Training Poses}} & \multicolumn{3}{c}{\textit{Novel Poses}} \\
        \cmidrule(r){3-5}  \cmidrule(lr){6-8}
        Pose & Avatar & \textbf{PSNR} & \textbf{SSIM} & \textbf{LPIPS} & \textbf{PSNR} & \textbf{SSIM} & \textbf{LPIPS}  \\ 
        \cmidrule(r){1-5}  \cmidrule(lr){6-8}
        \multicolumn{2}{c}{\textit{Better Together}} & 31.39 & 0.9586 & 0.1056 &  \multicolumn{3}{c}{-}  \\
        \midrule
        Ours & Ours & 31.53 & 0.9589 & 0.1041 & 31.61 & 0.9603 & 0.1043  \\
        Ours & HuGS~\cite{moreau2024human} & 31.37 & 0.9581 & 0.1003 & 31.53 & 0.9593 & 0.1015  \\
        DNA~\cite{cheng2023dna} & Ours & 30.09 & 0.9479 & 0.1271 & 29.18 & 0.9412 & 0.1314  \\ 
        DNA~\cite{cheng2023dna} & HuGS~\cite{moreau2024human} & 30.03 & 0.9475 & 0.1210 & 29.34 & 0.9416 & 0.1229  \\ \bottomrule
    \end{tabular}
    \end{adjustbox}
\end{table}

\paragraph{Per subject results} The results reported in Table~\ref{tab:DNA} of the main paper contain PSNR, SSIM and LPIPS averaged over evaluated subjects. In Table~\ref{tab:supp_DNA_subjects} we present the detailed results for each of the 5 evaluated subjects. Conclusions drawn on the main paper for this experiment remain valid for each subject.

\begin{figure*}[h]
\centering
\includegraphics[width=0.99\textwidth]{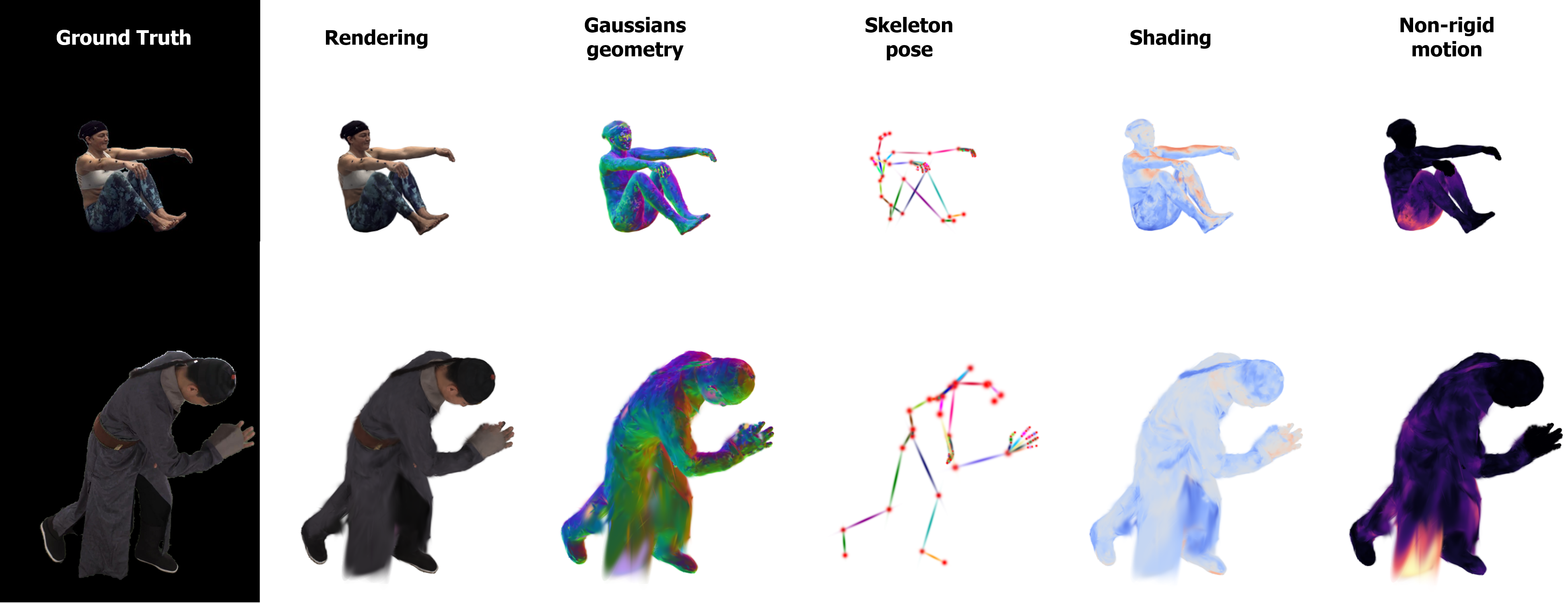}
\caption{\textbf{Visualization of reconstruction results of our method.}}
\label{fig:supp_reconstruction}
\end{figure*}

\paragraph{Visualization of reconstruction} We show in Figure~\ref{fig:supp_reconstruction} reconstruction results of our method on both MOYO and DNA-Rendering datasets. We observe that on the MOYO subject, where clothing is very close to the body, the non-rigid motion is used to fit body parts such as posterior and knees that are deformed in the sitting position (compared to the canonical pose which is standing in T-Pose). We also validate that shading succesfully captures the local lightning variations observed in the ground truth image. On the DNA Rendering subject, non-rigid motion is used to model deformation of loose clothing that is not tied to the skeleton.

\paragraph{On the importance of pose-dependent non-rigid motion} In the DNA-Rendering experiments described in the main paper we argue that ``our avatar is designed to rely on the pose signal only to explain details, which we observe to be important for robust pose estimation". We present here the experiment that brought us to this conclusion, where we use a different design for our body model. Instead of the Mesh MLP that computes non-rigid motion $m(\theta)$ and shading $s(\theta)$ from the body pose signal, we use a discrete implementation where non-rigid motion $m \in \mathbb{R}^{T \times V \times 3}$ and shading $s \in \mathbb{R}^{T \times V \times 1}$ are tensors learned explicitly for each timestep. This indeed gives more expressivity to the model that is not limited to the input pose signal to learn details, which is known as being insufficient~\cite{bagautdinov2021driving}. This can be explained intuitively: clothing deformation follow temporal dynamics and their position does not only depend on the current position but also on the previous motion. We performed the same experiment than in Table~\ref{tab:DNA} with the explicit (time-dependent) non-rigid motion and shading. 

First results where convincing: NVS results of Better Together on training poses is improved from 29.31db PSNR with the Mesh MLP to 30.14db PSNR. However, when training HuGS with the estimated parameters, quality degrades against results from Table~\ref{tab:DNA} : 28.31 and 26.96dB PSNR for training and novel poses respectively against 29.20dB and 27.84dB with parameters obtained from the Mesh MLP version.

The conclusion is that, despite a better photometric alignment of the renderings with the training poses, using time-dependent motion degrades the human pose estimation accuracy. The reason is that, even with an incorrect skeleton pose, the model is able to move the gaussians to the correct location. Is it not only bad for human pose estimation but also for novel pose animation because the skeleton is not correctly aligned with the mesh. This problem does not happen with the Mesh MLP because the driving signal is constrained to the skeleton pose only.

\section{Implementation details}

\paragraph{Anatomy consistent hand pose parameters} As explained in the main paper, we use a customized parameters space for hand pose with 22DoF per hand that constraints finger poses to anatomically plausible movements. Every finger joint is allowed to bend (closing hand movement) between 0 and 90 degrees, which is 15DoF. Then, the first joint of each finger can spread lateraly between -5 and 5 degrees (respectively -15 and 45 degrees for the thumb).
Finally, the thumb can also twist on his first joint and spread on the second, leading to 22 degrees of freedom. We learn 22 parameters which are processed by $sin$ activation function to be mapped to $[0,1]$ and then scaled to their pre-defined boundaries. We obtain Euler angles that are then decoded into SMPLx hand pose parameters by the anatomy aligned axis layer of ManoTorch~\cite{yang2021cpf}. 

\paragraph{Optimization}

We use the following hyperparameters during optimization. We first give the weighting of each regularization loss :

\begin{itemize}
    \item $\lambda_{normals} = 0.02$
    \item $\lambda_{m} = 0.2$
    \item $\lambda_{s} = 0.0005$
    \item $\lambda_{o} = 0.02$
    \item $\lambda_{rot} = 0.001$
\end{itemize}

And the learning rates of parameters:
\begin{itemize}
    \item Motion MLPs : 1e-4
    \item Mesh MLP : 1e-4
    \item Gaussians MLP : 1e-4
    \item Body shape $\beta$ : 1e-3 
    \item Vertices offset $\Delta v$ : 5e-4 to 5e-5 (linear decay) 
    \item Barycentric coordinates : 0.1
    \item Latent features $f$ : 0.0025
\end{itemize}

\section{Data preparation}

In this section, we describe the procedure to use our method on newly collected data, which corresponds to what we used for the MOYO dataset~\cite{tripathi2023ipman}. It is the same procedure that is commonly used to create animatable avatars, excluding the pose parameters estimation which is solved directly by our method. Note that for DNA-Rendering dataset, keypoints and masks are already provided so we just used this data. We assume that camera poses are known which is usually the case in the lightstage setups we consider. Before applying the method described in the main paper, we perform the following pre-processing steps:

\paragraph{Keypoints detection and triangulation} We use Sapiens~\cite{khirodkar2025sapiens}, a recent human foundation model to extract 2D keypoints with confidence on each video frame for each camera. More specifically, we used the \textit{Sapiens 2b} model with \textit{COCO wholebody} keypoints convention. Then, for each timestep, we triangulate 2D into 3D using camera parameters with the Direct Linear Transforms (DLT) algorithm, using keypoints confidence as weighting. We obtain 3D keypoints that are used in the pretraining step of our method for motion MLPs initialization (but not for the main optimization step with photometric supervision). Removing entirely the dependency to keypoints would be an interesting research direction to improve our method.

\paragraph{Foreground masks extraction} We use SAM2~\cite{ravi2024sam2} to extract foreground masks of the human, using detected keypoints from the body as point prompts for the mask. We also tried to use Sapiens segmentation model for this purpose but observed sub optimal results. Foreground masks are used to remove the background in captured images such that the rendering task consists only in reconstructing the human. While this step is necessary for now, we observed that masks are never perfect and that imprecise segmentation introduces visual artefacts in the human avatar model. One potential solution to avoid this problem could be to reconstruct the entire images, including background. It could be done by having a pre-computed 3DGS model of the environment where the motion is captured and combine both background and human model during the rendering of images.

\paragraph{Image cropping (optional)} To save memory space and computation, we crop images to remove as much background pixels where the human is not observed as possible. For each camera, we define a cropping area as the smallest bounding box that contains all the segmentation masks of the human observed during a sequence. All images from the same camera are cropped with the same bounding box and the camera intrinsics parameters are adjusted accordingly.

\end{document}